\documentclass{sig-alternate-05-2015}
\usepackage{array}
\usepackage[none]{hyphenat}
\usepackage{subfigure}
\usepackage{amsmath}
\newcolumntype{L}{>{\centering\arraybackslash}m{1.4cm}}

\usepackage{graphicx}

\usepackage{etoolbox}

\global\copyrightetc{Presented at 1st ACM SIGKDD Workshop on Machine Learning for Prognostics and Health Management, San Francisco, CA, USA, 2016. Copyright \the\copyrtyr\ Tata Consultancy Services Ltd. }

\begin{document}
\sloppy

\title{Multi-Sensor Prognostics using an Unsupervised Health Index based on LSTM Encoder-Decoder\vspace{-2ex}}
\author{
\normalsize Pankaj Malhotra, Vishnu TV, Anusha Ramakrishnan\\
\normalsize Gaurangi Anand, Lovekesh Vig, Puneet Agarwal, Gautam Shroff\\
       \affaddr{\normalsize TCS Research, New Delhi, India}\\
       \email{\small\{malhotra.pankaj, vishnu.tv, anusha.ramakrishnan\}@tcs.com}\\ 
       \email{\small\{gaurangi.anand, lovekesh.vig, puneet.a, gautam.shroff\}@tcs.com}\\ 
\vspace{-2ex}}
\maketitle
\begin{abstract}
Many approaches for estimation of Remaining Useful Life (RUL) of a machine, using its operational sensor data, make assumptions about how a system degrades or a fault evolves, e.g., exponential degradation. However, in many domains degradation may not follow a pattern. We propose a Long Short Term Memory based Encoder-Decoder (LSTM-ED) scheme to obtain an unsupervised health index (HI) for a system using multi-sensor time-series data. LSTM-ED is trained to reconstruct the time-series corresponding to healthy state of a system. The reconstruction error is used to compute HI which is then used for RUL estimation. We evaluate our approach on publicly available Turbofan Engine and Milling Machine datasets. We also present results on a real-world industry dataset from a pulverizer mill where we find significant correlation between LSTM-ED based HI and maintenance costs.
\end{abstract}

\section{Introduction}
Industrial Internet has given rise to availability of sensor data from numerous machines belonging to various domains such as agriculture, energy, manufacturing etc. These sensor readings can indicate health of the machines. This has led to increased business desire to perform maintenance of these machines based on their condition rather than following the current industry practice of time-based maintenance. It has also been shown that condition-based maintenance can lead to significant financial savings. Such goals can be achieved by building models for prediction of remaining useful life (RUL) of the machines, based on their sensor readings.

Traditional approach for RUL prediction is based on an assumption that the health degradation curves (drawn w.r.t. time) follow specific shape such as exponential or linear. Under this assumption we can build a model for health index (HI) prediction, as a function of sensor readings. Extrapolation of HI is used for prediction of RUL \cite{p:rulclipper,mosallam2014data,mosallam2015component}. However, we observed that such assumptions do not hold in the real-world datasets, making the problem harder to solve. Some of the important challenges in solving the prognostics problem are: i) health degradation curve may not necessarily follow a fixed shape, ii) time to reach same level of degradation by machines of same specifications is often different, iii) each instance has a slightly different initial health or wear, iv) sensor readings if available are noisy, v) sensor data till end-of-life is not easily available because in practice periodic maintenance is performed.

Apart from the health index (HI) based approach as described above, mathematical models of the underlying physical system, fault propagation models and conventional reliability models have also been used for RUL estimation \cite{cadini2009model,myotyri2006application}. Data-driven models which use readings of sensors carrying degradation or wear information such as vibration in a bearing have been effectively used to build RUL estimation models \cite{qiu2003robust,p:rulclipper,wang2001fault}. Typically, sensor readings over the entire operational life of multiple instances of a system from start till failure are used to obtain common degradation behavior trends or to build models of how a system degrades by estimating health in terms of HI. Any new instance is then compared with these trends and the most similar trends are used to estimate the RUL \cite{p:similarity_wang2008}. 

LSTM networks are recurrent neural network models that have been successfully used for many sequence learning and temporal modeling tasks \cite{graves2009novel,anand2015deep} such as handwriting recognition, speech recognition, sentiment analysis, and customer behavior prediction. A variant of LSTM networks, LSTM encoder-decoder (LSTM-ED) model has been successfully used for sequence-to-sequence learning tasks \cite{p:seq2seq,p:seq2seqNIPS2014,p:tensorflowEncDec} like machine translation, natural language generation and reconstruction, parsing, and image captioning. LSTM-ED works as follows: An LSTM-based encoder is used to map a multivariate input sequence to a fixed-dimensional vector representation. The decoder is another LSTM network which uses this vector representation to produce the target sequence. We provide further details on LSTM-ED in Sections \ref{ssec:lstm} and \ref{ssec:lstm-ed}.

LSTM Encoder-decoder based approaches have been proposed for anomaly detection \cite{p:icmlLSTM-AD,marchi2015novel}. These approaches learn a model to reconstruct the normal data (e.g. when machine is in perfect health) such that the learned model could reconstruct the subsequences which belong to normal behavior. The learnt model leads to high reconstruction error for anomalous or novel subsequences, since it has not seen such data during training. Based on similar ideas, we use Long Short-Term Memory \cite{hochreiter1997long} Encoder-Decoder (LSTM-ED) for RUL estimation. In this paper, we propose an unsupervised technique to obtain a health index (HI) for a system using multi-sensor time-series data, which does not make any assumption on the shape of the degradation curve. We use LSTM-ED to learn a model of normal behavior of a system, which is trained to reconstruct multivariate time-series corresponding to normal behavior. The reconstruction error at a point in a time-series is used to compute HI at that point. In this paper, we show that:
\begin{itemize}
 \item \textit{LSTM-ED based HI learnt in an unsupervised manner is able to capture the degradation in a system: the HI decreases as the system degrades.}
 \item \textit{LSTM-ED based HI can be used to learn a model for RUL estimation instead of relying on domain knowledge, or exponential/linear degradation assumption, while achieving comparable performance.}
\end{itemize}

The rest of the paper is organized as follows: We formally introduce the problem and provide an overview of our approach in Section \ref{sec:Approach}. 
In Section \ref{sec:LR-HI}, we describe Linear Regression (LR) based approach to estimate the health index and discuss commonly used assumptions to obtain these estimates.
In Section \ref{sec:LSTM-ED}, we describe how LSTM-ED can be used to learn the LR model without relying on domain knowledge or mathematical models for degradation evolution. In Section \ref{sec:RUL_Estimation}, we explain how we use the HI curves of train instances and a new test instance to estimate the RUL of the new instance. We provide details of experiments and results on three datasets in Section \ref{sec:Experiments}. Finally, we conclude with a discussion in Section \ref{sec:Discussion}, after a summary of related work in Section \ref{sec:Related_Work}.

\section{Approach Overview}\label{sec:Approach}
We consider the scenario where historical instances of a system with multi-sensor data readings till end-of-life are available. The goal is to estimate the RUL of a currently operational instance of the system for which multi-sensor data is available till current time-instance.

More formally, we consider a set of train instances $U$ of a system. For each instance $u \in U$, we consider a multivariate time-series of sensor readings $X^{(u)}=[\mathbf{x}^{(u)}_{1}\: \mathbf{x}^{(u)}_{2}\: ... \: \mathbf{x}^{(u)}_{L^{(u)}}]$ with $L^{(u)}$ cycles where the last cycle corresponds to the end-of-life, each point $\mathbf{x}^{(u)}_{t} \in \mathbf{R}^m$ is an $m$-dimensional vector corresponding to readings for $m$ sensors at time-instance $t$. The sensor data is z-normalized such that the sensor reading $x_{tj}^{(u)}$ at time $t$ for $j$th sensor for instance $u$ is transformed to $\frac{x_{tj}^{(u)}-\mu_j}{\sigma_j}$, where $\mu_j$ and $\sigma_j$ are mean and standard deviation for the $j$th sensor's readings over all cycles from all instances in $U$. (Note: When multiple modes of normal operation exist, each point can normalized based on the $\mu_j$ and $\sigma_j$ for that mode of operation, as suggested in \cite{thesis:tsbp}.) A subsequence of length $l$ for time-series $X^{(u)}$ starting from time instance $t$ is denoted by $X^{(u)}(t,l)=[x^{(u)}_{t}\:x^{(u)}_{t+1}\: ...\: x^{(u)}_{t+l-1}]$ with $1\leq t \leq L^{(u)}-l+1$. 

In many real-world multi-sensor data, sensors are correlated. As in \cite{thesis:tsbp,mosallam2014data,mosallam2015component}, we use Principal Components Analysis to obtain derived sensors from the normalized sensor data with reduced linear correlations between them. The multivariate time-series for the derived sensors is represented as
$Z^{(u)}=[\mathbf{z}^{(u)}_{1}\: \mathbf{z}^{(u)}_{2}\: ...\: \mathbf{z}^{(u)}_{L^{(u)}}]$, where $\mathbf{z}^{(u)}_{t}\in \mathbf{R}^p$, $p$ is the number of principal components considered (The best value of $p$ can be obtained using a validation set).

For a new instance $u^*$ with sensor readings $X^{(u^*)}$ over $L^{(u^*)}$ cycles, the goal is to estimate the remaining useful life $R^{(u^*)}$ in terms of number of cycles, given the time-series data for train instances \{$X^{(u)}: u \in U$\}. We describe our approach assuming same operating regime for the entire life of the system. The approach can be easily extended to multiple operating regimes scenario by treating data for each regime separately (similar to \cite{thesis:tsbp,lam2014enhanced}), as described in one of our case studies on milling machine dataset in Section \ref{ssec:milling}.

\begin{figure*}[htp]
 \centering
\includegraphics[width=0.85\textwidth]{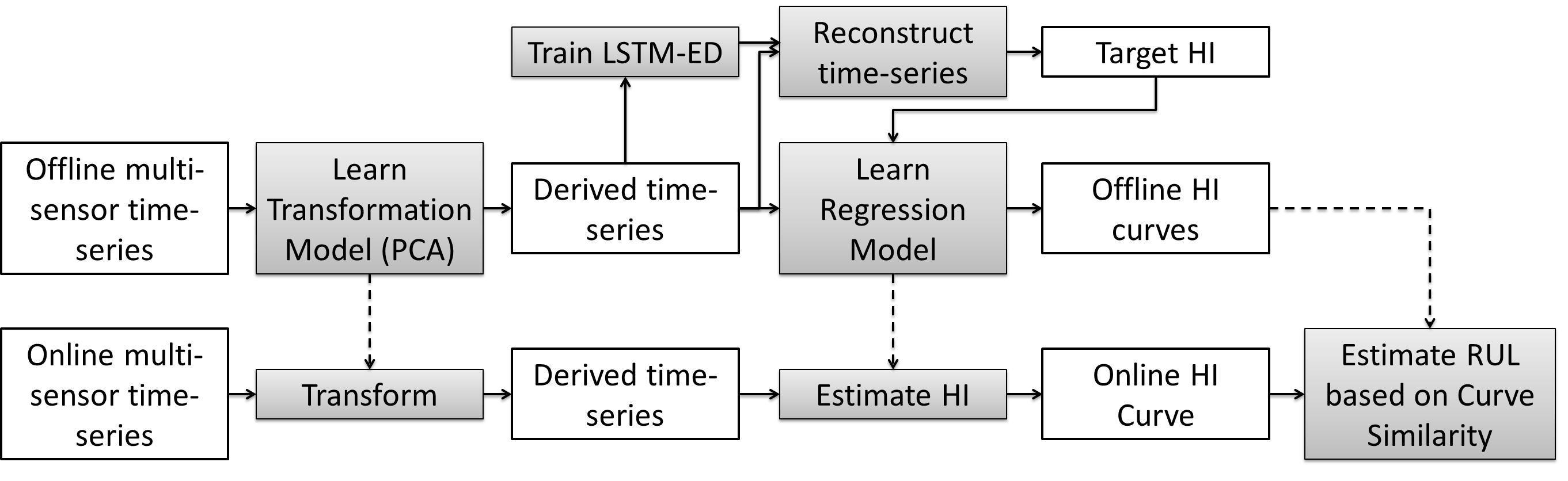}
\vspace{-8pt}
\caption{RUL estimation steps using unsupervised HI based on LSTM-ED.}
\vspace{-8pt}
\end{figure*}

\section{Linear Regression based Health Index Estimation}\label{sec:LR-HI}
Let $H^{(u)}=[h^{(u)}_{1}\:h^{(u)}_{2}\: ...\: h^{(u)}_{L^{(u)}}]$ represent the HI curve $H^{(u)}$ for instance $u$, where each point $h^{(u)}_{t} \in \mathbf{R}$, $L^{(u)}$ is the total number of cycles. We assume $0\leq h^{(u)}_{t}\leq 1$, s.t. when $u$ is in perfect health $h^{(u)}_{t}=1$, and when $u$ performs below an acceptable level (e.g. instance is about to fail), $h^{(u)}_{t}=0$.

Our goal is to construct a mapping $f_{\boldsymbol{\theta}}: \mathbf{z}^{(u)}_{t} \rightarrow h^{(u)}_{t}$ 
s.t. 
\begin{equation}\label{eq:lr}
f_{\boldsymbol{\theta}}(\mathbf{z}^{(u)}_{t})=\boldsymbol{\theta}^T\mathbf{z}^{(u)}_{t}+\theta_0
\end{equation}
where $\boldsymbol{\theta}\in \mathbf{R}^p$, $\theta_0 \in \mathbf{R}$, which computes HI $h^{(u)}_{t}$ from the derived sensor readings $\mathbf{z}^{(u)}_{t}$ at time $t$ for instance $u$. 
Given the \textit{target HI curves} for the training instances, the parameters $\boldsymbol{\theta}$ and $\theta_0$ are estimated using Ordinary Least Squares method.

\subsection{Domain-specific target HI curves}
The parameters $\boldsymbol{\theta}$ and $\theta_0$ of the above mentioned Linear Regression (LR) model  (Eq. \ref{eq:lr}) are usually estimated by assuming a mathematical form for the target $H^{(u)}$, with an exponential function being the most common and successfully employed target HI curve (e.g. \cite{croarkin2006nist,saxena2008damage,p:similarity_wang2008,p:rulclipper,camci2016comparison}), which assumes the HI at time $t$ for instance $u$ as
\begin{equation}\label{eq:expAssumption}
 h^{(u)}_t=1-exp\left(\frac{log(\beta).(L^{(u)}-t)}{(1-\beta).L^{(u)}}\right), t \in [\beta.L^{(u)},(1-\beta).L^{(u)}].
\end{equation}
$0<\beta<1$. The starting and ending $\beta$ fraction of cycles are assigned values of $1$ and $0$, respectively.

Another possible assumption is: assume target HI values of $1$ and $0$ for data corresponding to healthy condition and failure conditions, respectively. Unlike the exponential HI curve which uses the entire time-series of sensor readings, the sensor readings corresponding to only these points are used to learn the regression model (e.g. \cite{wang2012generic}).

The estimates $\boldsymbol{\hat{\theta}}$ and $\hat{\theta_0}$ based on target HI curves for train instances are used to obtain the \textit{final HI curves} $H^{(u)}$ for all the train instances and a new test instance for which RUL is to be estimated. The HI curves thus obtained are used to estimate the RUL for the test instance based on similarity of train and test HI curves, as described later in Section \ref{sec:RUL_Estimation}.

\section{LSTM-ED based target HI curve}\label{sec:LSTM-ED}
We learn an LSTM-ED model to reconstruct the time-series of the train instances during normal operation. For example, any subsequence corresponding to starting few cycles when the system can be assumed to be in healthy state can be used to learn the model. The reconstruction model is then used to reconstruct all the subsequences for all train instances and the pointwise reconstruction error is used to obtain a target HI curve for each instance. We briefly describe LSTM unit and LSTM-ED based reconstruction model, and then explain how the reconstruction errors obtained from this model are used to obtain the target HI curves.

\subsection{LSTM unit}\label{ssec:lstm}
An LSTM unit is a recurrent unit that uses the input $z_t$, the hidden state activation $a_{t-1}$, and memory cell activation $c_{t-1}$ to compute the hidden state activation $a_t$ at time $t$. It uses a combination of a memory cell $c$ and three types of gates: input gate $i$, forget gate $f$, and output gate $o$ to decide if the input needs to be remembered (using input gate), when the previous memory needs to be retained (forget gate), and when the memory content needs to be output (using output gate).

Many variants and extensions to the original LSTM unit as introduced in \cite{hochreiter1997long} exist. We use the one as described in \cite{zaremba2014recurrent}.
Consider $T_{n_1,n_2}:\mathbf{R}^{n_1} \rightarrow \mathbf{R}^{n_2}$ is an affine transform of the form $\mathbf{z}\mapsto \mathbf{Wz}+\mathbf{b}$ for matrix $\mathbf{W}$ and vector $\mathbf{b}$ of appropriate dimensions.
The values for input gate $i$, forget gate $f$, output gate $o$, hidden state $a$, and cell activation $c$ at time $t$ are computed using the current input $z_t$, the previous hidden state $a_{t-1}$, and memory cell value $c_{t-1}$ as given by Eqs. \ref{eq:lstm1}-\ref{eq:lstm3}.

\begin{equation}\label{eq:lstm1}
  \left(\begin{aligned}
   i_t\\
   f_t\\
   o_t\\
   g_t
  \end{aligned}\right)=\left(\begin{aligned}
   \sigma\quad\\
   \sigma\quad\\
   \sigma\quad\\
   tanh\\
  \end{aligned}\right)T_{m+n,4n}
  \left(\begin{aligned}
    z_{t}\:\:\\
    a_{t-1}\\
    \end{aligned}\right)
\end{equation}

Here $\sigma(z)=\frac{1}{1+e^{-z}}$ and $tanh(z)=2\sigma(2z)-1$. The operations $\sigma$ and $tanh$ are applied elementwise. The four equations from the above simplifed matrix notation read as: $i_t=\sigma(W_1z_t+W_2a_{t-1}+b_i)$, etc. Here, $x_t\in \mathbf{R}^m$, and all others $i_t,f_t,o_t,g_t,a_t,c_t\in \mathbf{R}^n$.
\begin{equation}\label{eq:lstm2}
 c_t=f_tc_{t-1}+i_tg_t
\end{equation}
\begin{equation}\label{eq:lstm3}
a_t=o_{t}tanh(c_t)
\end{equation}


\begin{figure}
 \centering
\includegraphics[width=\columnwidth]{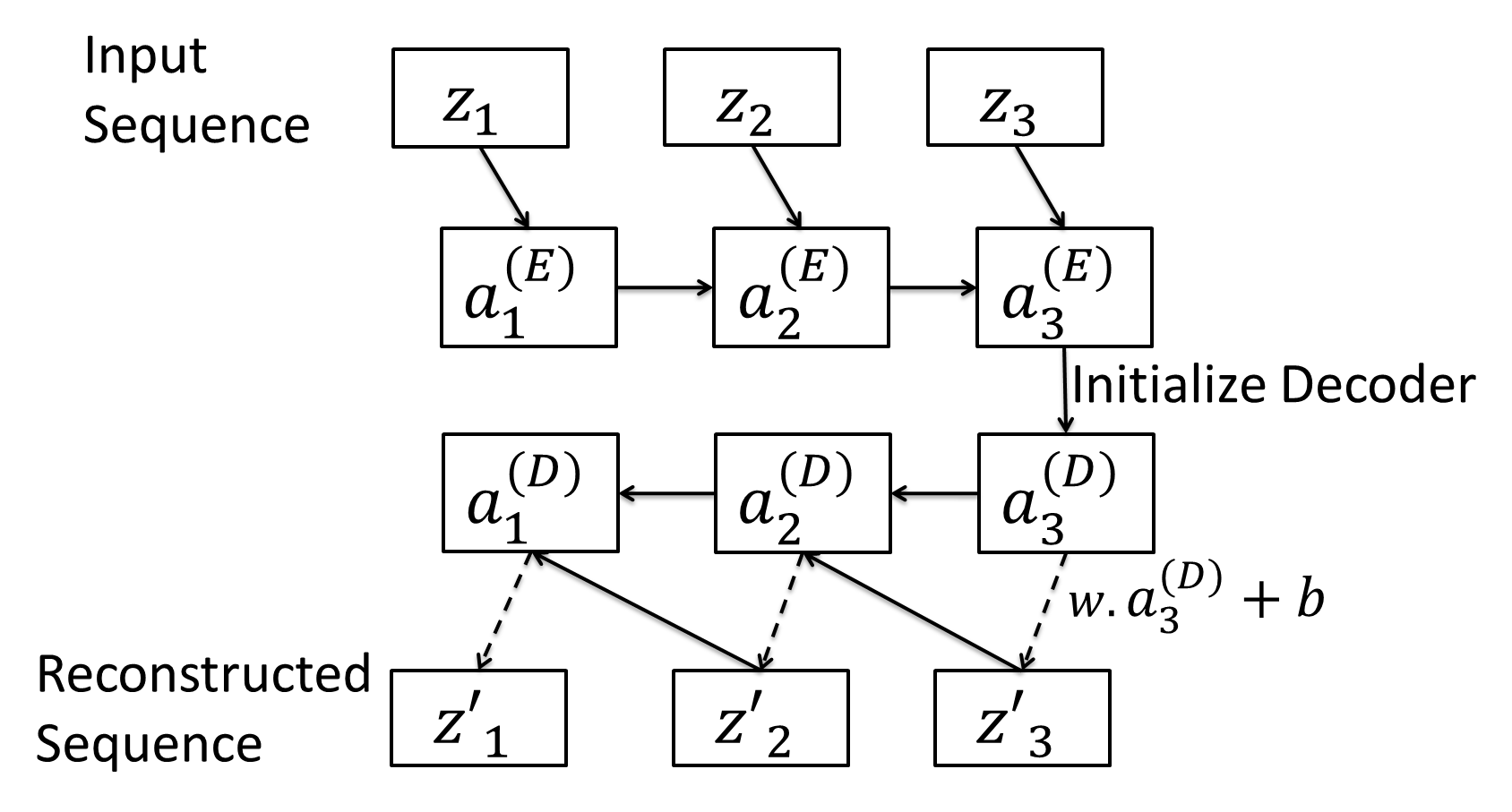}
\caption{\label{fig:EncDec}LSTM-ED inference steps for input $\{\mathbf{z}_{1},\mathbf{z}_{2}, \mathbf{z}_{3}\}$ to predict $\{\mathbf{z'}_{1},\mathbf{z'}_{2}, \mathbf{z'}_{3}\}$}
\end{figure}

\subsection{Reconstruction Model}\label{ssec:lstm-ed}
We consider sliding windows to obtain $L-l+1$ subsequences for a train instance with $L$ cycles. LSTM-ED is trained to reconstruct the normal (healthy) subsequences of length $l$ from all the training instances. The LSTM encoder learns a fixed length vector representation of the input time-series and the LSTM decoder uses this representation to reconstruct the time-series using the current hidden state and the value predicted at the previous time-step. Given a time-series $Z=[\mathbf{z}_{1}\:\mathbf{z}_{2}\:...\:\mathbf{z}_{l}]$, $\mathbf{a}^{(E)}_{t}$ is the hidden state of encoder at time $t$ for each $t \in \{1,2,...,l\}$, where $\mathbf{a}^{(E)}_{t} \in \mathbf{R}^c$, $c$ is the number of LSTM units in the hidden layer of the encoder. The encoder and decoder are jointly trained to reconstruct the time-series in reverse order (similar to \cite{p:seq2seqNIPS2014}), i.e., the target time-series is $[\mathbf{z}_{l}\:\mathbf{z}_{l-1}\:...\:\mathbf{z}_{1}]$. (Note: We consider derived sensors from PCA s.t. $m=p$ and $n=c$ in Eq. \ref{eq:lstm1}.)

Fig. \ref{fig:EncDec} depicts the inference steps in an LSTM-ED reconstruction model for a toy sequence with $l=3$.
The value $\mathbf{x}_{t}$ at time instance $t$ and the hidden state $\mathbf{a}^{(E)}_{t-1}$ of the encoder at time $t-1$ are used to obtain the hidden state $\mathbf{a}^{(E)}_{t}$ of the encoder at time $t$. The hidden state $\mathbf{a}^{(E)}_{l}$ of the encoder at the end of the input sequence is used as the initial state $\mathbf{a}^{(D)}_{l}$ of the decoder s.t. $\mathbf{a}^{(D)}_{l}=\mathbf{a}^{(E)}_{l}$. A linear layer with weight matrix $\mathbf{w}$ of size $c\times m$ and bias vector $\mathbf{b} \in \mathbf{R}^m$ on top of the decoder is used to compute $\textbf{z}'_{t}=\mathbf{w^Ta}^{(D)}_{t}+\mathbf{b}$. During training, the decoder uses $\mathbf{z}_{t}$ as input to obtain the state $\mathbf{a}^{(D)}_{t-1}$, and then predict $\mathbf{z'}_{t-1}$ corresponding to target $\mathbf{z}_{t-1}$. During inference, the predicted value $\mathbf{z'}_{t}$ is input to the decoder to obtain $\mathbf{a}^{(D)}_{t-1}$ and predict $\mathbf{z'}_{t-1}$.
The reconstruction error $e_t^{(u)}$ for a point $\mathbf{z}_t^{(u)}$ is given by:
\begin{equation}\label{eq:lstm-ed-err}
 e_t^{(u)}=\|\mathbf{z}^{(u)}_{t}-\mathbf{z'}^{(u)}_{t}\|
\end{equation}
The model is trained to minimize the objective $E=\sum_{u \in U}\sum_{t=1}^{l}(e_t^{(u)})^2$. 
It is to be noted that for training, only the subsequences which correspond to perfect health of an instance are considered. For most cases, the first few operational cycles can be assumed to correspond to healthy state for any instance.

\subsection{Reconstruction Error based Target HI}\label{ssec:unsupervised_targetHI}
A point $\mathbf{z}_t$ in a time-series $Z$ is part of multiple overlapping subsequences, and is therefore predicted by multiple subsequences $Z(j,l)$ corresponding to $j=t-l+1, t-l+2,...,t$. Hence, each point in the original time-series for a train instance is predicted as many times as the number of subsequences it is part of ($l$ times for each point except for points $\mathbf{z}_t$ with $t<l$ or $t>L-l$ which are predicted fewer number of times). An average of all the predictions for a point is taken to be final prediction for that point. The difference in actual and predicted values for a point is used as an unnormalized HI for that point. 

Error $e_t^{(u)}$ is normalized to obtain the \textit{target HI} $h_t^{(u)}$ as:
\begin{equation}\label{eq:targetHI}
 h_t^{(u)}=\frac{e_{M}^{(u)}-e_t^{(u)}}{e_{M}^{(u)}-e_{m}^{(u)}}
\end{equation}

where $e_{M}^{(u)}$ and $e_{m}^{(u)}$ are the maximum and minimum values of reconstruction error for instance $u$ over $t=1\: 2\:...\: L^{(u)}$, respectively. The target HI values thus obtained for all train instances are used to obtain the estimates $\boldsymbol{\hat{\theta}}$ and $\hat{\theta_0}$ (see Eq. \ref{eq:lr}). Apart from $e_t^{(u)}$, we also consider $(e_t^{(u)})^2$ to obtain target HI values for our experiments in Section \ref{sec:Experiments} such that large reconstruction errors imply much smaller HI value.

\section{RUL estimation using HI curve matching}\label{sec:RUL_Estimation}
\begin{figure}
\centering
\includegraphics[width=\columnwidth]{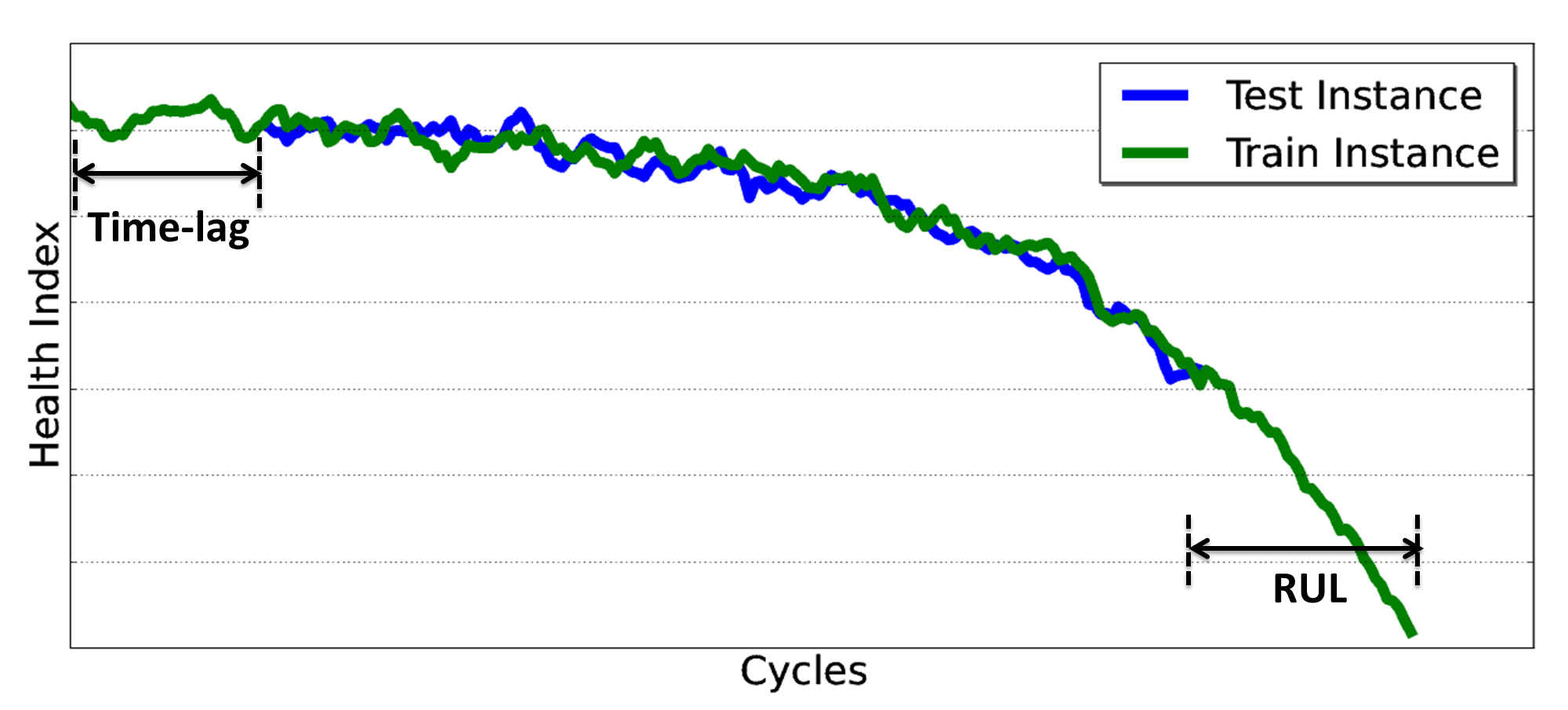}
\vspace{-4pt}
\caption{\label{fig:time-lag}Example of RUL estimation using HI curve matching taken from Turbofan Engine dataset (refer  Section \ref{ssec:engine_data})}
\vspace{-4pt}
\end{figure}
Similar to \cite{thesis:tsbp,p:similarity_wang2008}, the HI curve for a test instance $u^*$ is compared to the HI curves of all the train instances $u\in U$. The test instance and train instance may take different number of cycles to reach the same degradation level (HI value). Fig. \ref{fig:time-lag} shows a sample scenario where HI curve for a test instance is matched with HI curve for a train instance by varying the time-lag. The time-lag which corresponds to minimum Euclidean distance between the HI curves of the train and test instance is shown. For a given time-lag, the number of remaining cycles for the train instance after the last cycle of the test instance gives the RUL estimate for the test instance.
Let $u^*$ be a test instance and $u$ be a train instance.
Similar to \cite{p:rulclipper,thesis:2datasets,p:rnnRUL1}, we take into account the following scenarios for curve matching based RUL estimation:
\newline
1) \textit{Varying initial health across instances}: The initial health of an instance varies depending on various factors such as the inherent inconsistencies in the manufacturing process. We assume initial health to be close to $1$. In order to ensure this, the HI values for an instance are divided by the average of its first few HI values  (e.g. first 5\% cycles). Also, while comparing HI curves $H^{(u^*)}$ and $H^{(u)}$, we allow for a time-lag $t$ such that the HI values of $u^*$ may be close to the HI values of $H^{(u)}(t,L^{(u^*)})$ at time $t$ such that $t\leq \tau$ (see Eqs. \ref{eq:sim}-\ref{eq:rul_estimate}). This takes care of instance specific variances in degree of initial wear and degradation evolution.
\newline
2) \textit{Multiple time-lags with high similarity}: The HI curve $H^{(u^*)}$ may have high similarity with $H^{(u)}(t,L^{(u*)})$ for multiple values of time-lag $t$. We consider multiple RUL estimates for $u^*$ based on total life of $u$, rather than considering only the RUL estimate corresponding to the time-lag $t$ with minimum Euclidean distance between the curves $H^{(u^*)}$ and $H^{(u)}(t,L^{(u^*)})$. The multiple RUL estimates corresponding to each time-lag  are assigned weights proportional to the similarity of the curves to get the final RUL estimate (see Eq. \ref{eq:rul_estimate}).
\newline
3) \textit{Non-monotonic HI}: Due to inherent noise in sensor readings, HI curves obtained using LR are non-monotonic. To reduce the noise in the estimates of HI, we use moving average smoothening.
\newline
4) \textit{Maximum value of RUL estimate}: When an instance is in very good health or has been operational for few cycles, estimating RUL is difficult. We limit the maximum RUL estimate for any test instance to $R_{max}$. Also, the maximum RUL estimate for the instance $u^*$ based on HI curve comparison with instance $u$ is limited by $L^{(u)}-L^{(u^*)}$. This implies that the maximum RUL estimate for any test instance $u$ will be such that the total length $\hat{R}^{(u^*)}+L^{(u^*)}\leq L_{max}$, where $L_{max}$ is the maximum length for any training instance available. Fewer the number of cycles available for a test instance, more difficult it becomes to estimate the RUL.

We define similarity between HI curves of test instance $u^*$ and train instance $u$ with time-lag $t$ as: \begin{equation}\label{eq:sim}
 s(u^*,u,t)=exp(-d^2(u^*,u,t)/\lambda)
\end{equation}
where, 
\begin{equation}
 d^2(u^*,u,t)=\frac{1}{L^{(u^*)}}\sum_{i=1}^{L^{(u^*)}}{(h_i^{(u^*)}-h_{i+t}^{(u)})^2}
\end{equation}

is the squared Euclidean distance between $H^{(u^*)}(1,L^{(u^*)})$ and $H^{(u)}(t,L^{(u^*)})$, and $\lambda>0$, $t\in \{1,2,...,\tau\}$, $t+L^{(u^*)} \leq L^{(u)}$. Here, $\lambda$ controls the notion of similarity: a small value of $\lambda$ would imply large difference in $s$ even when $d$ is not large. The RUL estimate for $u^*$ based on the HI curve for $u$ and for time-lag $t$ is given by $\hat{R}^{(u^*)}(u,t)=L^{(u)}-L^{(u^*)}-t$.

The estimate $\hat{R}^{(u^*)}(u,t)$ is assigned a weight of $s(u^*,u,t)$ such that the weighted average estimate $\hat{R}^{(u^*)}$ for $R^{(u^*)}$ is given by 
\begin{equation}\label{eq:rul_estimate}
 \hat{R}^{(u^*)}=\frac{\sum{s(u^*,u,t).\hat{R}^{(u^*)}(u,t)}}{\sum{s(u^*,u,t)}}
\end{equation}
where the summation is over only those combinations of $u$ and $t$ which satisfy $s(u^*,u,t)\geq \alpha.s_{max}$, where $s_{max}=max_{u\in U,t \in \{1\:...\:\tau\}}\{s(u^*,u,t)\}$, $0\leq \alpha \leq 1$. 

It is to be noted that the parameter $\alpha$ decides the number of RUL estimates $\hat{R}^{(u^*)}(u,t)$ to be considered to get the final RUL estimate $\hat{R}^{(u^*)}$.
Also, variance of the RUL estimates $\hat{R}^{(u^*)}(u,t)$ considered for computing $\hat{R}^{(u^*)}$ can be used as a measure of confidence in the prediction, which is useful in practical applications (for example, see Section \ref{sssec:aircraft_observ}). During the initial stages of an instance's usage, when it is in good health and a fault has still not appeared, estimating RUL is tough, as it is difficult to know beforehand how exactly the fault would evolve over time once it appears.

\section{Experimental Evaluation}\label{sec:Experiments}
We evaluate our approach on two publicly available datasets: C-MAPSS Turbofan Engine Dataset \cite{saxena2008damage} and Milling Machine Dataset \cite{milling_dataset}, and a real world dataset from a pulverizer mill. For the first two datasets, the ground truth in terms of the RUL is known, and we use RUL estimation performance metrics to measure efficacy of our algorithm (see Section \ref{ssec:performance_metrics}). The Pulverizer mill undergoes repair on timely basis (around one year), and therefore ground truth in terms of actual RUL is not available. We therefore draw comparison between health index and the cost of maintenance of the mills.

For the first two datasets, we use different target HI curves for learning the LR model (refer Section \ref{sec:LR-HI}): \textit{LR-Lin} and \textit{LR-Exp} models assume linear and exponential form for the target HI curves, respectively. \textit{LR-ED$_1$} and \textit{LR-ED$_2$} use normalized reconstruction error and normalized squared-reconstruction error as target HI (refer Section \ref{ssec:unsupervised_targetHI}), respectively. The target HI values for LR-Exp are obtained using Eq. \ref{eq:expAssumption} with $\beta=5\%$ as suggested in \cite{p:similarity_wang2008,thesis:tsbp,p:rulclipper}.

\subsection{Performance metrics considered}\label{ssec:performance_metrics}
Several metrics have been proposed for evaluating the performance of prognostics models \cite{saxena2008metrics}. We measure the performance in terms of Timeliness Score (S), Accuracy (A), Mean Absolute Error (MAE), Mean Squared Error (MSE), and Mean Absolute Percentage Error (MAPE$_1$ and MAPE$_2$)  as mentioned in Eqs. \ref{eq:timeliness}-\ref{eq:mape-2}, respectively.
For test instance $u^*$, the error $\Delta^{(u^*)} =\hat{R}^{(u^*)} - {R}^{(u^*)}$ between the estimated RUL ($\hat{R}^{(u^*)}$) and actual RUL ($R^{(u^*)}$). The score $S$ used to measure the performance of a model is given by:
\begin{equation}\label{eq:timeliness}
 S=\sum^N_{u^*=1} (exp({\gamma.|\Delta^{(u^*)}|})-1)
\end{equation}
where $\gamma=1/\tau_1$ if $\Delta^{(u^*)}<0$, else $\gamma=1/\tau_2$. Usually, $\tau_1 > \tau_2$ such that late predictions are penalized more compared to early predictions. The lower the value of $S$, the better is the performance.

\begin{equation}\label{eq:accuracy}
 A=\frac{100}{N}\sum^N_{u^*=1} I(\Delta^{(u^*)}) 
\end{equation}
where $I(\Delta^{(u^*)})=1 \: if \: \Delta^{(u^*)}\in [-\tau_1,\tau_2]$, else  $I(\Delta^{(u^*)})=0$, $\tau_1>0, \tau_2>0$. 

\begin{equation}\label{eq:mae}
MAE=\frac{1}{N}\sum^N_{u^*=1}|\Delta^{(u^*)}|, \:MSE=\frac{1}{N}\sum^N_{u^*=1}(\Delta^{(u^*)})^2
\end{equation}

\begin{equation}
 MAPE_1=\frac{100}{N}\sum^N_{u^*=1}\frac{|\Delta^{(u^*)}|}{R^{(u^*)}}
\end{equation}

\begin{equation}\label{eq:mape-2}
 MAPE_2=\frac{100}{N}\sum^N_{u^*=1}\frac{|\Delta^{(u^*)}|}{R^{(u^*)}+L^{(u^*)}}
\end{equation}

A prediction is considered a false positive (FP) if $\Delta^{(u^*)}<-\tau_1$, and false negative (FN) if $\Delta^{(u^*)}>\tau_2$.
\subsection{C-MAPSS Turbofan Engine Dataset}\label{ssec:engine_data}
\begin{figure}
\centering
\includegraphics[scale=0.27]{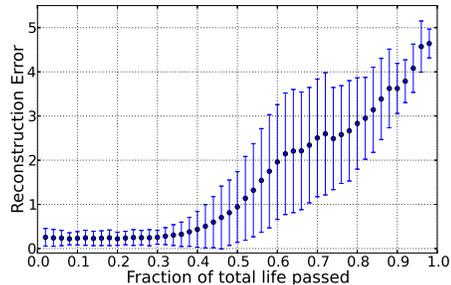}
\caption{\label{fig:error-1sigma}Turbofan Engine: Average pointwise reconstruction error based on LSTM-ED w.r.t. fraction of total life passed.}
\vspace{-6pt}
\end{figure}
We consider the first dataset from the simulated turbofan engine data \cite{saxena2008damage} (NASA Ames Prognostics Data Repository). The dataset contains readings for $24$ sensors ($3$ operation setting sensors, $21$ dependent sensors) for $100$ engines till a failure threshold is achieved, i.e., till end-of-life in $train\_FD001.txt$. Similar data is provided for $100$ test engines in $test\_FD001.txt$ where the time-series for engines are pruned some time prior to failure. The task is to predict RUL for these $100$ engines. The actual RUL values are provided in $RUL\_FD001.txt$. There are a total of $20631$ cycles for training engines, and $13096$ cycles for test engines. Each engine has a different degree of initial wear.
We use $\tau_1=13, \tau_2=10$ as proposed in \cite{saxena2008damage}.

\begin{figure*}[htp]
 \centering
 \subfigure[LSTM-ED\label{fig:err_hist_ED_raw}]{\includegraphics[width=0.22\textwidth]{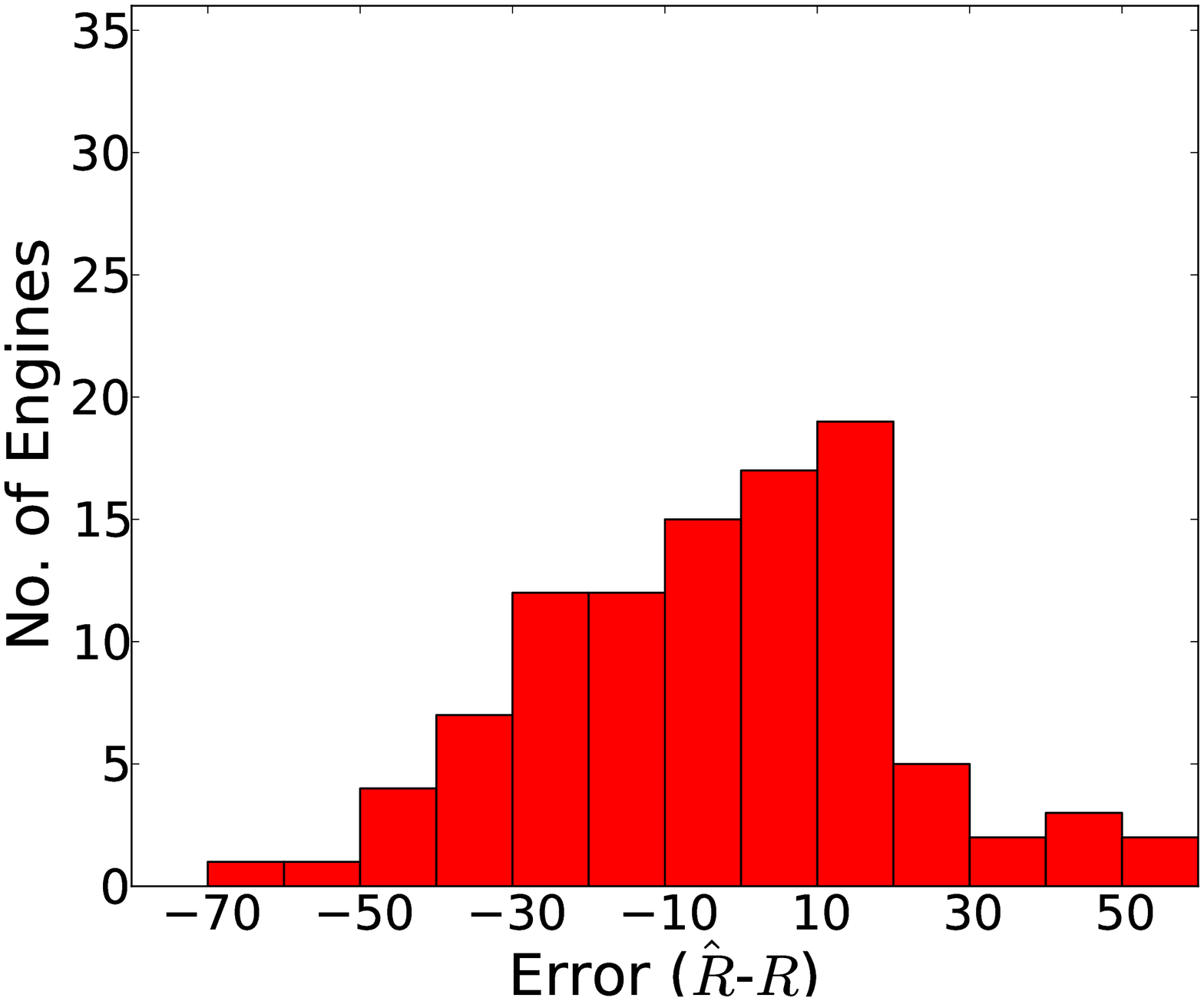}}
 \subfigure[LR-Exp]{\includegraphics[width=0.22\textwidth]{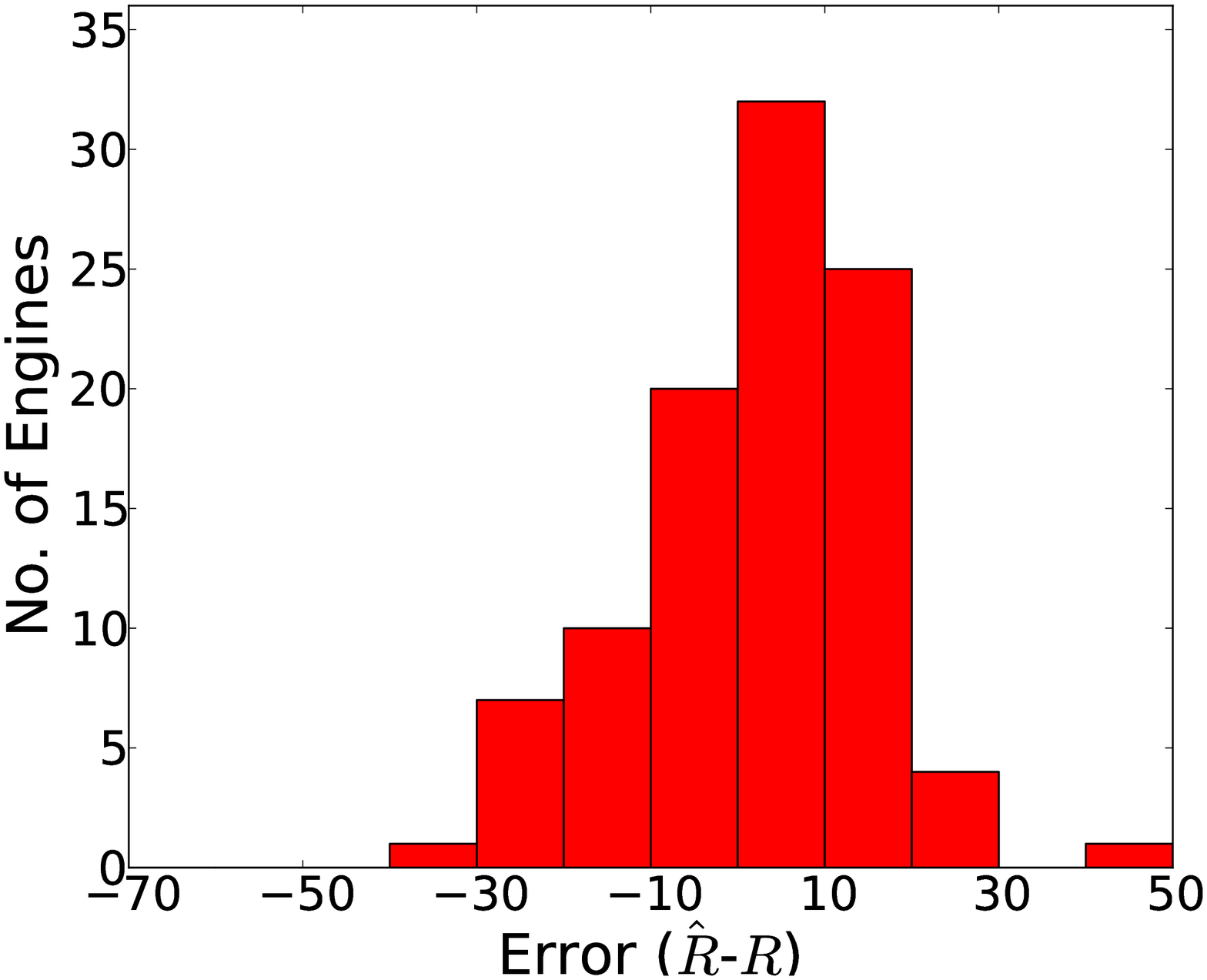}}
 \subfigure[LR-ED$_1$]{\includegraphics[width=0.22\textwidth]{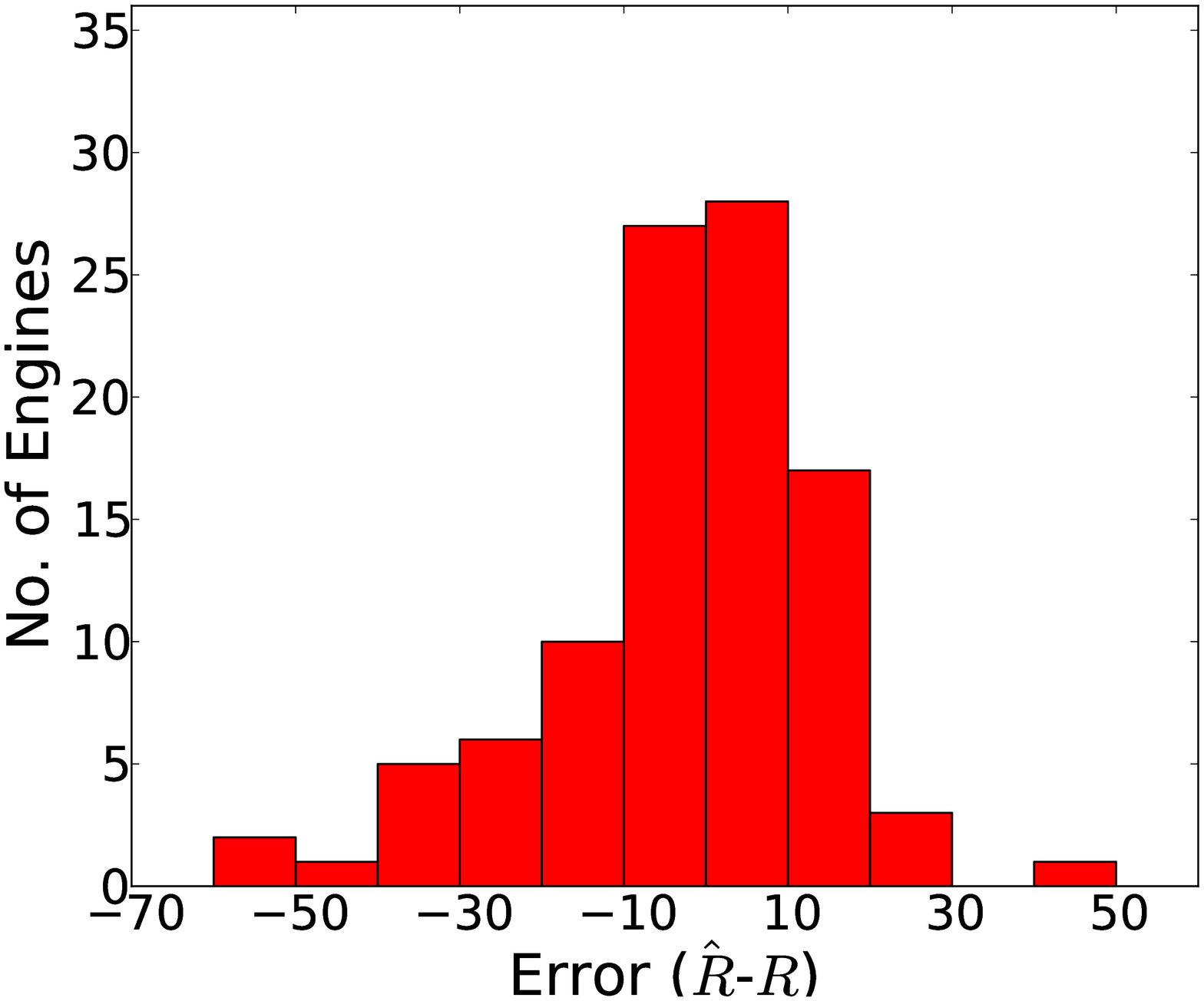}}
 \subfigure[LR-ED$_2$]{\includegraphics[width=0.22\textwidth]{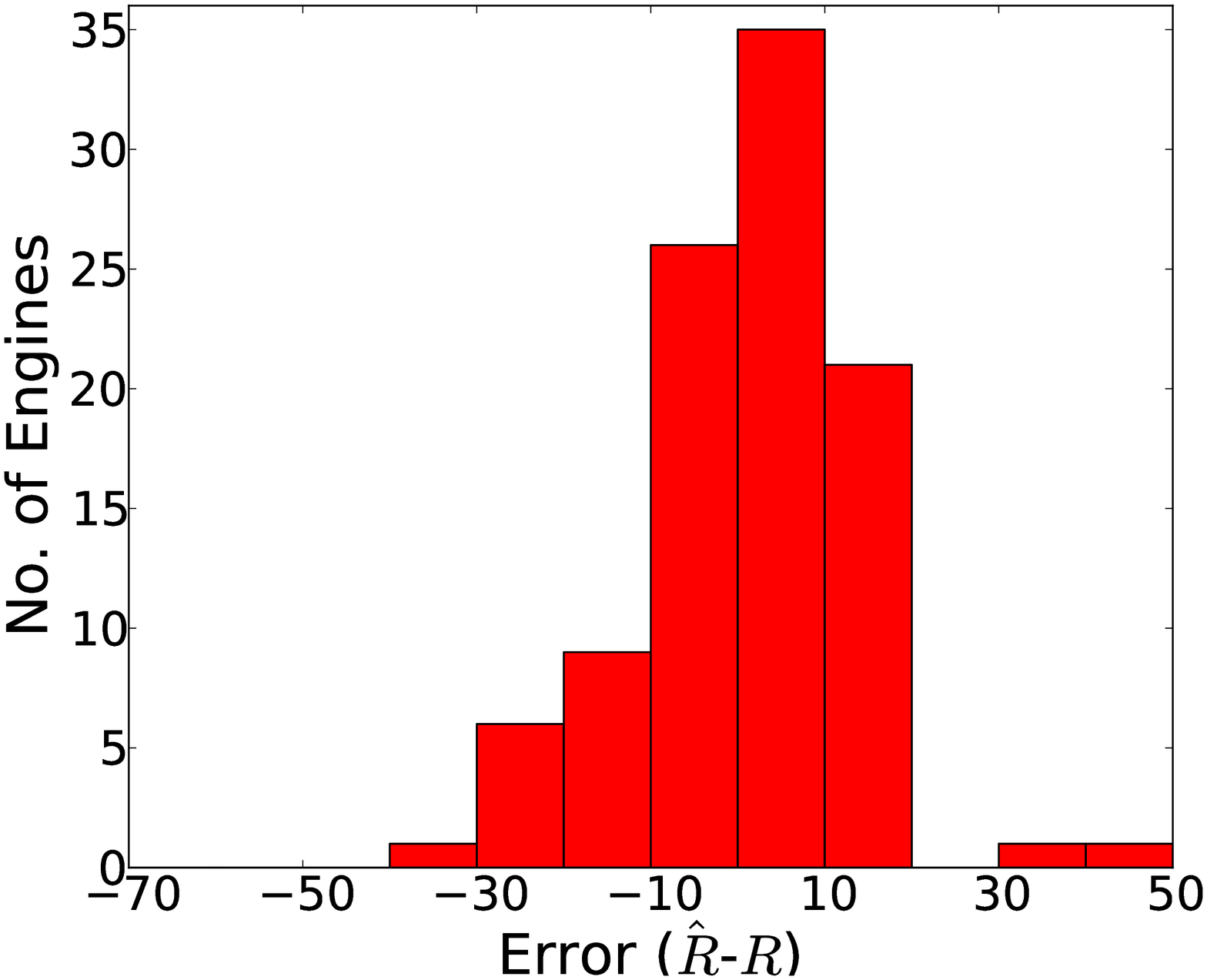}} 
 \caption{\label{fig:err_hist}Turbofan Engine: Histograms of prediction errors for Turbofan Engine dataset.}
 \vspace{-10pt}
\end{figure*}

\begin{figure*}[htp]
 \centering
 \subfigure[RUL estimates given by LR-Exp, LR-ED$_1$, and LR-ED$_2$\label{fig:predictions}]{\includegraphics[width=0.65\textwidth]{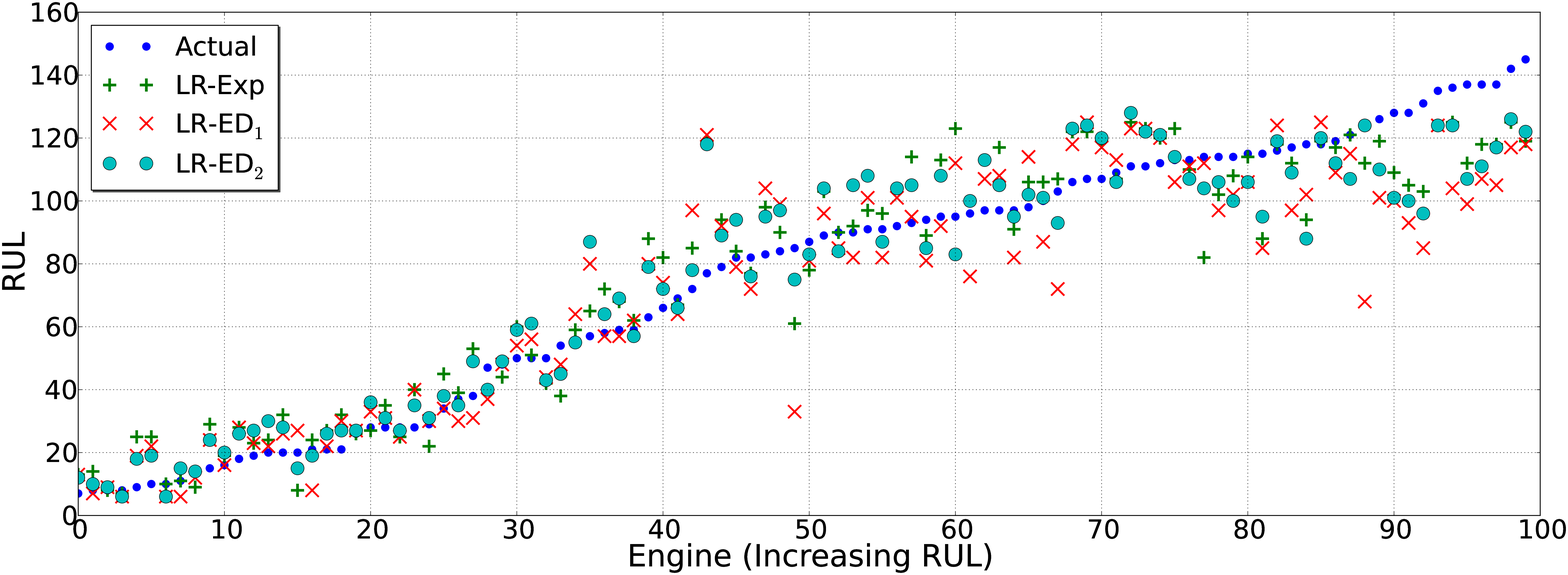}}
 \subfigure[Standard deviation, max-min, and absolute error w.r.t HI at last cycle\label{fig:confidence_wrt_HI}]{\includegraphics[scale=0.24]{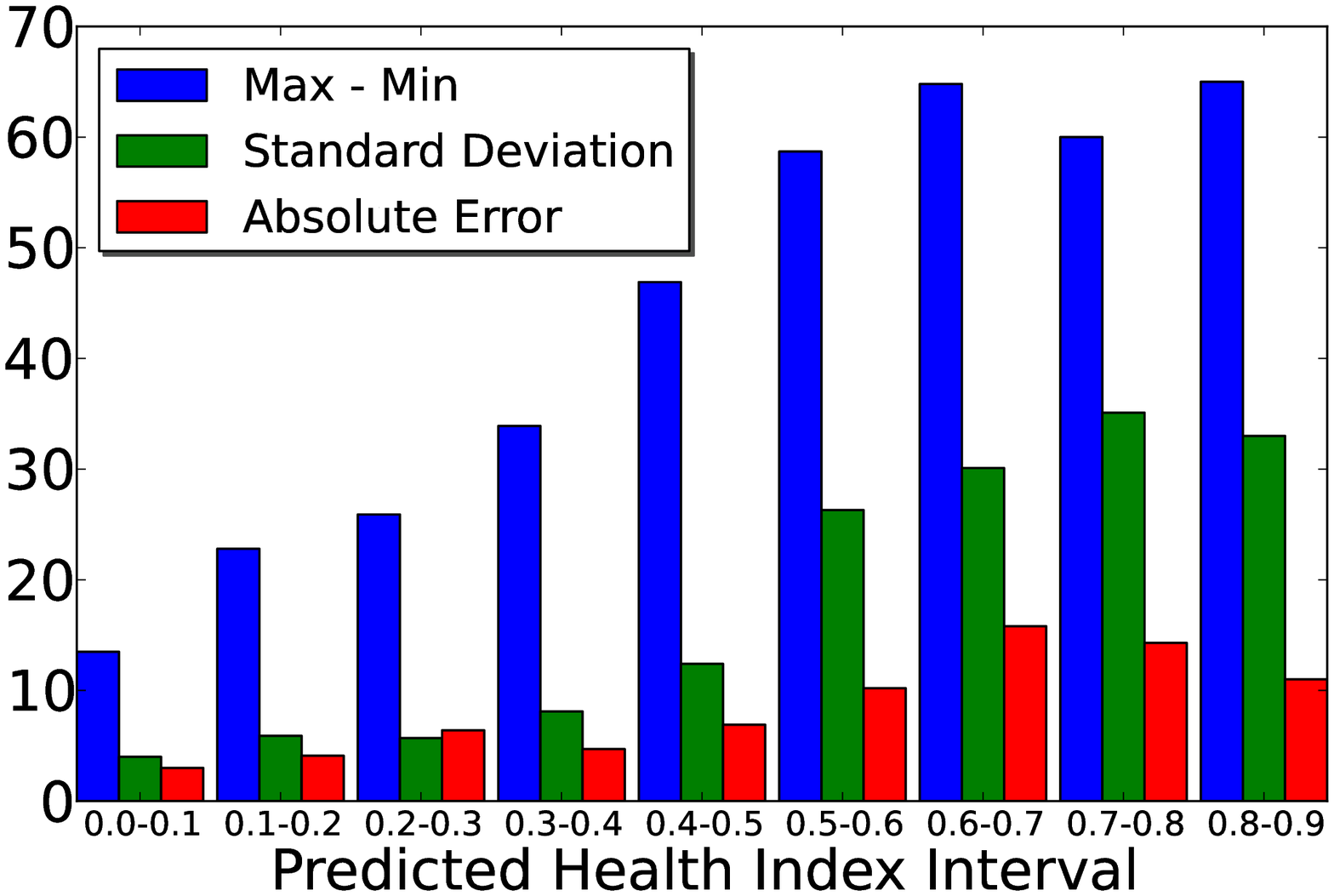}}
 \caption{Turbofan Engine: (a) RUL estimates for all 100 engines, (b) Absolute error w.r.t. HI at last cycle}
 \vspace{-10pt}
\end{figure*}

\subsubsection{Model learning and parameter selection}
We randomly select $80$ engines for training the LSTM-ED model and estimating parameters $\boldsymbol{\theta}$ and $\theta_0$ of the LR model (refer Eq. \ref{eq:lr}). The remaining $20$ training instances are used for selecting the parameters. The trajectories for these $20$ engines are randomly truncated at five different locations s.t. five different cases are obtained from each instance. Minimum truncation is $20\%$ of the total life and maximum truncation is $96\%$. For training LSTM-ED, only the first subsequence of length $l$ for each of the selected $80$ engines is used. 
The parameters number of principal components $p$, the number of LSTM units in the hidden layers of encoder and decoder $c$, window/subsequence length $l$, maximum allowed time-lag $\tau$, similarity threshold $\alpha$ (Eq. \ref{eq:rul_estimate}), maximum predicted RUL $R_{max}$, and parameter $\lambda$ (Eq. \ref{eq:sim}) are estimated using grid search to minimize $S$ on the validation set. The parameters obtained for the best model (LR-ED$_2$) are $p=3$, $c=30$, $l=20$, $\tau=40$, $\alpha=0.87$, $R_{max}=125$, and $\lambda=0.0005$.

\subsubsection{Results and Observations}\label{sssec:aircraft_observ}
\textbf{LSTM-ED based Unsupervised HI}: 
Fig. \ref{fig:error-1sigma} shows the average pointwise reconstruction error (refer Eq. \ref{eq:lstm-ed-err}) given by the model LSTM-ED which uses the pointwise reconstruction error as an unnormalized measure of health (higher the reconstruction error, poorer the health) of all the $100$ test engines w.r.t. percentage of life passed (for derived sensor sequences with $p=3$ as used for model LR-ED$_1$ and LR-ED$_2$). During initial stages of an engine's life, the average reconstruction error is small. \textit{As the number of cycles passed increases, the reconstruction error increases. This suggests that reconstruction error can be used as an indicator of health of a machine.} Fig. 5(a) and Table \ref{tab:aircraftPerf2} suggest that RUL estimates given by HI from LSTM-ED are fairly accurate. 
On the other hand, the 1-sigma bars in Fig. \ref{fig:error-1sigma} also suggest that the reconstruction error at a given point in time (percentage of total life passed) varies significantly from engine to engine. 

\textbf{Performance comparison}: Table \ref{tab:aircraftPerf2} and Fig. 5 show the performance of the four models LSTM-ED (without using linear regression), LR-Exp, LR-ED$_1$, and LR-ED$_2$. We found that LR-ED$_2$ performs significantly better compared to the other three models.  LR-ED$_2$ is better than the LR-Exp model which uses domain knowledge in the form of exponential degradation assumption. 
We also provide comparison with RULCLIPPER (RC) \cite{p:rulclipper} which (to the best of our knowledge) has the best performance in terms of timeliness $S$, accuracy $A$, MAE, and MSE \cite{ramasso2014performance} reported in the literature\footnote{For comparison with some of the other benchmarks readily available in literature, see Table \ref{tab:aircraftPerf1}. It is to be noted that the comparison is not exhaustive as a survey of approaches for the turbofan engine dataset since \cite{ramasso2014performance} is not available.} on the turbofan dataset considered and four other turbofan engine datasets (\textit{Note: }Unlike RC, we learn the parameters of the model on a validation set rather than test set.) RC relies on the exponential assumption to estimate a HI polygon and uses intersection of areas of polygons of train and test engines as a measure of similarity to estimate RUL (similar to Eq. \ref{eq:rul_estimate}, see \cite{p:rulclipper} for details). \textit{The results show that LR-ED$_2$ gives performance comparable to RC without relying on the domain-knowledge based exponential assumption.}

The worst predicted test instance for LR-Exp, LR-ED$_1$ and LR-ED$_2$ contributes $23\%$, $17\%$, and $23\%$, respectively, to the timeliness score $S$. For LR-Exp and LR-ED$_2$ it is nearly $1/4^{th}$ of the total score, and suggests that for other $99$ test engines the timeliness score $S$ is very good.

\begin{table}
\centering
\resizebox{\columnwidth}{!}{
\begin{tabular}{|c|c|c|c|c|c|} \hline
&\textbf{LSTM-ED} & \textbf{LR-Exp} & \textbf{LR-ED$_1$} & \textbf{LR-ED$_2$} & \textbf{RC}\\ \hline
\textbf{S} & 1263 & 280 & 477 & 256 & \textbf{216}\\ \hline
\textbf{A(\%)} & 36 & 60 & 65 & \textbf{67} & \textbf{67}\\ \hline
\textbf{MAE} & 18 & \textbf{10} & 12 & \textbf{10} & \textbf{10}\\ \hline
\textbf{MSE} & 546 & 177 & 288 & \textbf{164} & 176\\ \hline
\textbf{MAPE$_1$(\%)} & 39 & 21 & 20 & \textbf{18} & 20\\ \hline
\textbf{MAPE$_2$(\%)} & 9 & 5.2 & 5.9 & \textbf{5.0} & NR\\ \hline
\textbf{FPR(\%)} & 34 & \textbf{13} & 19 & \textbf{13} & 56\\ \hline
\textbf{FNR(\%)} & 30 & 27 & \textbf{16} & 20 & 44\\ \hline
\end{tabular}
}
\caption{\label{tab:aircraftPerf2}Turbofan Engine: Performance comparison}
 \vspace{-10pt}
\end{table}

\textbf{HI at last cycle and RUL estimation error}: Fig. \ref{fig:predictions} shows the actual and estimated RULs for LR-Exp, LR-ED$_1$, and LR-ED$_2$. For all the models, we observe that as the actual RUL increases, the error in predicted values increases. Let $R^{(u^*)}_{all}$ denote the set of all the RUL estimates $\hat{R}^{(u^*)}(u,t)$ considered to obtain the final RUL estimate $\hat{R}^{(u^*)}$ (see Eq. \ref{eq:rul_estimate}). Fig. \ref{fig:confidence_wrt_HI} shows the average values of absolute error, standard deviation of the elements in $R^{(u^*)}_{all}$, and the difference of the maximum and the minimum value of the elements in $R^{(u^*)}_{all}$ w.r.t. HI value at last cycle. It suggests that when an instance is close to failure, i.e., HI at last cycle is low, RUL estimate is very accurate with low standard deviation of the elements in $R^{(u^*)}_{all}$. On the other hand, when an instance is in good health, i.e., when HI at last cycle is close to 1, the error in RUL estimate is high, and the elements in $R^{(u^*)}_{all}$ have high standard deviation.

\subsection{Milling Machine Dataset}\label{ssec:milling}
\begin{figure*}[htp]
 \centering
 \subfigure[Recon. Error Material-1]{\includegraphics[width=0.48\columnwidth]{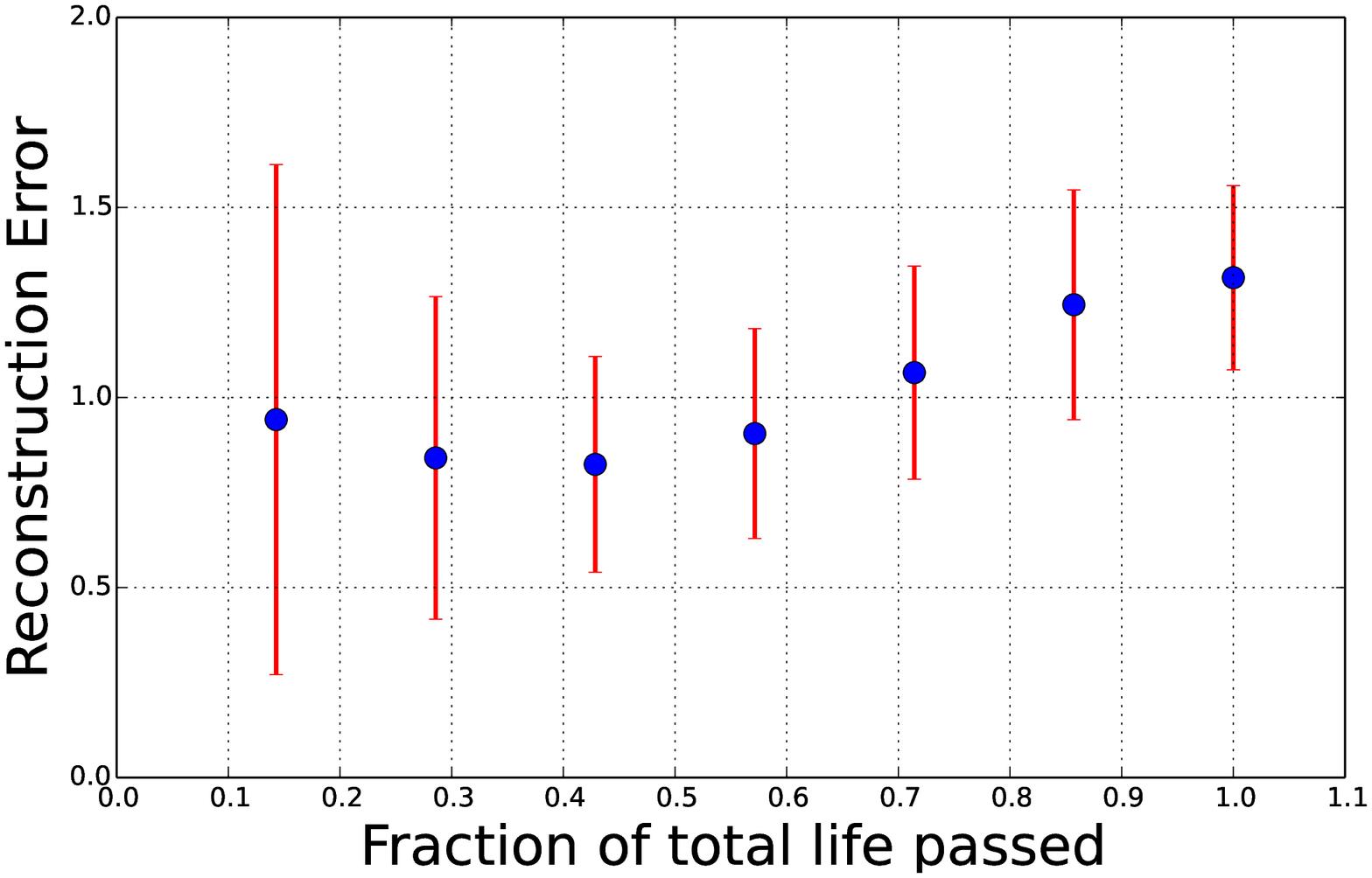}}
 \subfigure[PCA$_1$ Material-1\label{fig:err_hist_ED_raw}]{\includegraphics[width=0.24\textwidth]{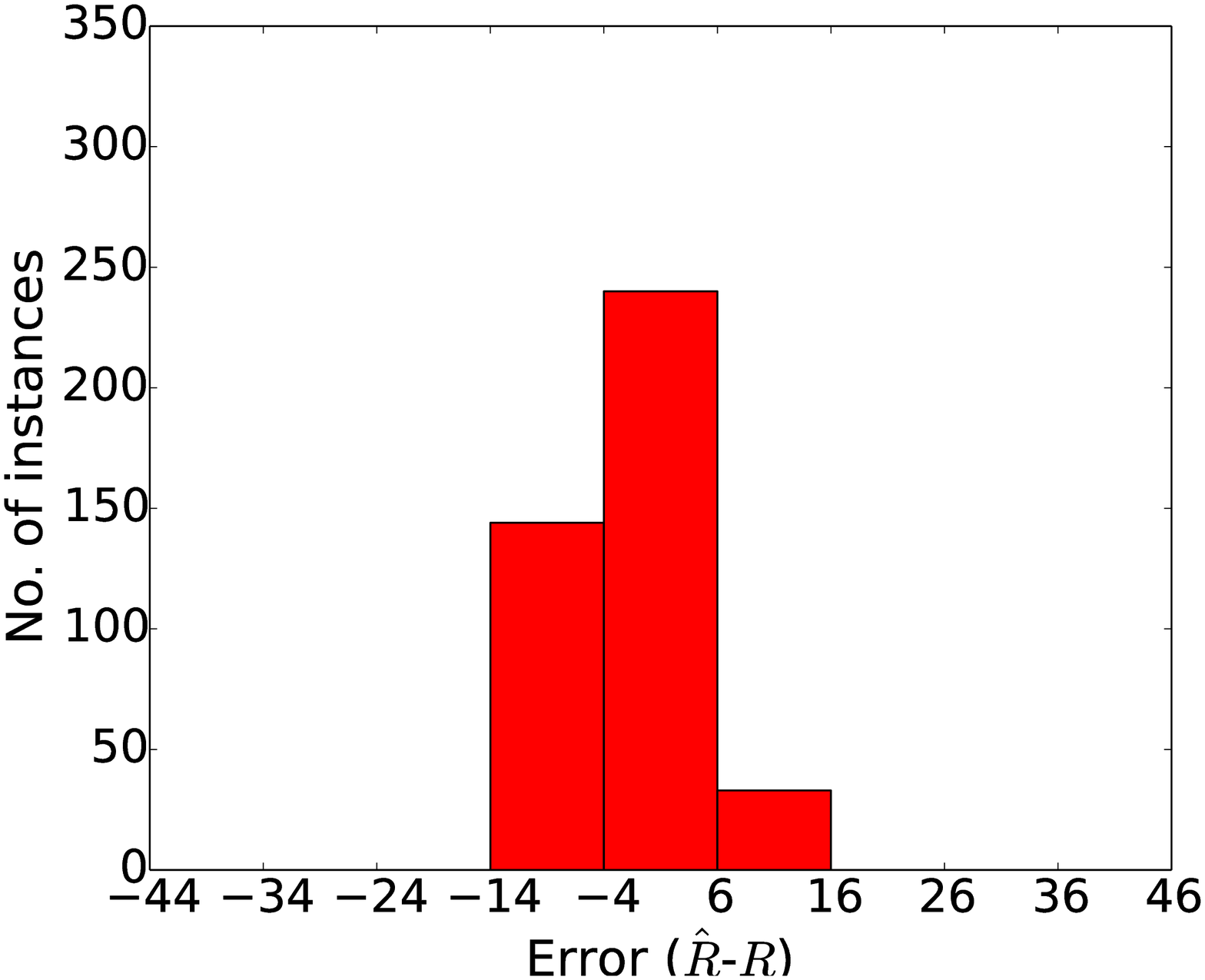}}
 \subfigure[LR-ED$_1$ Material-1]{\includegraphics[width=0.24\textwidth]{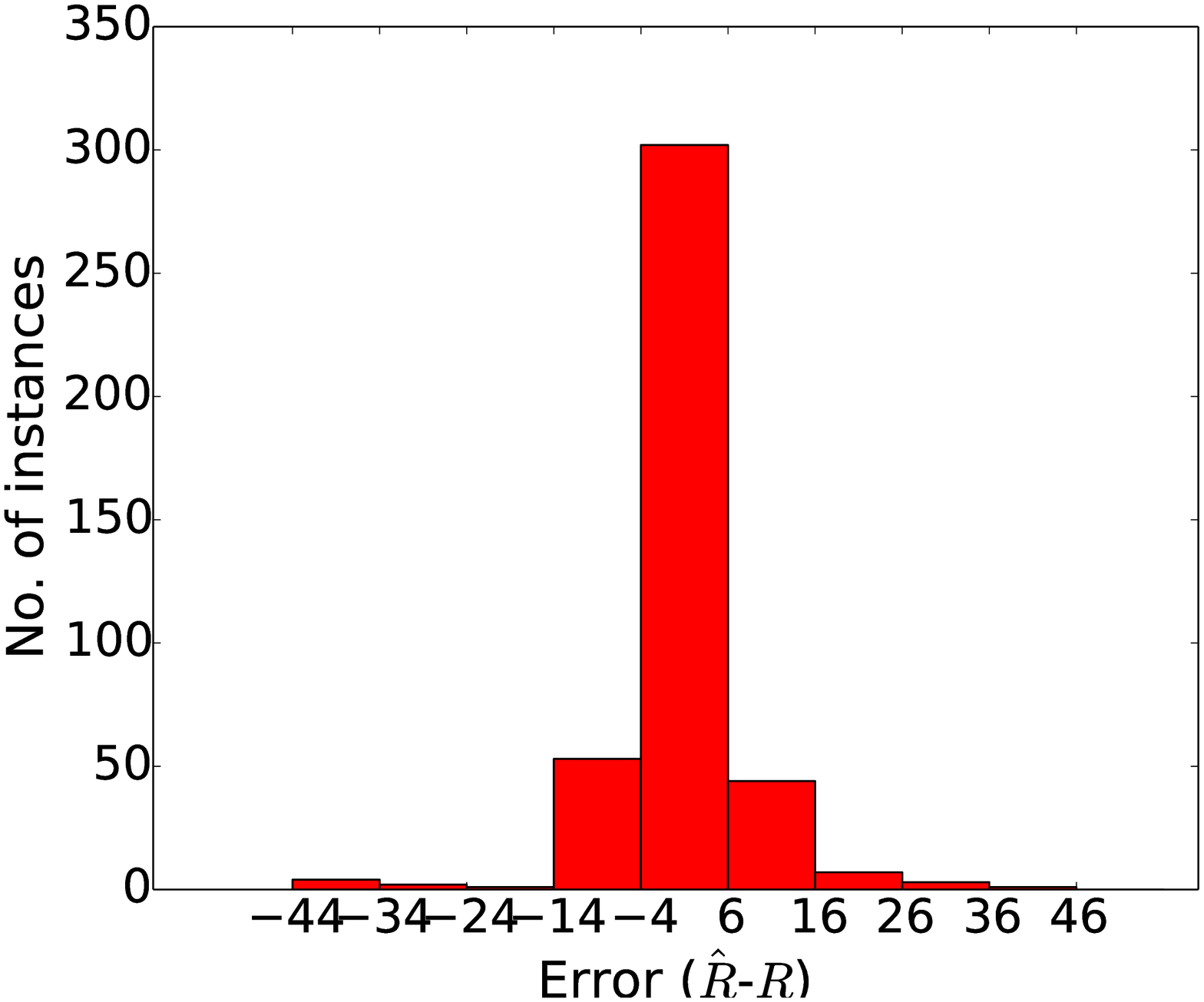}}
 \subfigure[LR-ED$_2$ Material-1]{\includegraphics[width=0.24\textwidth]{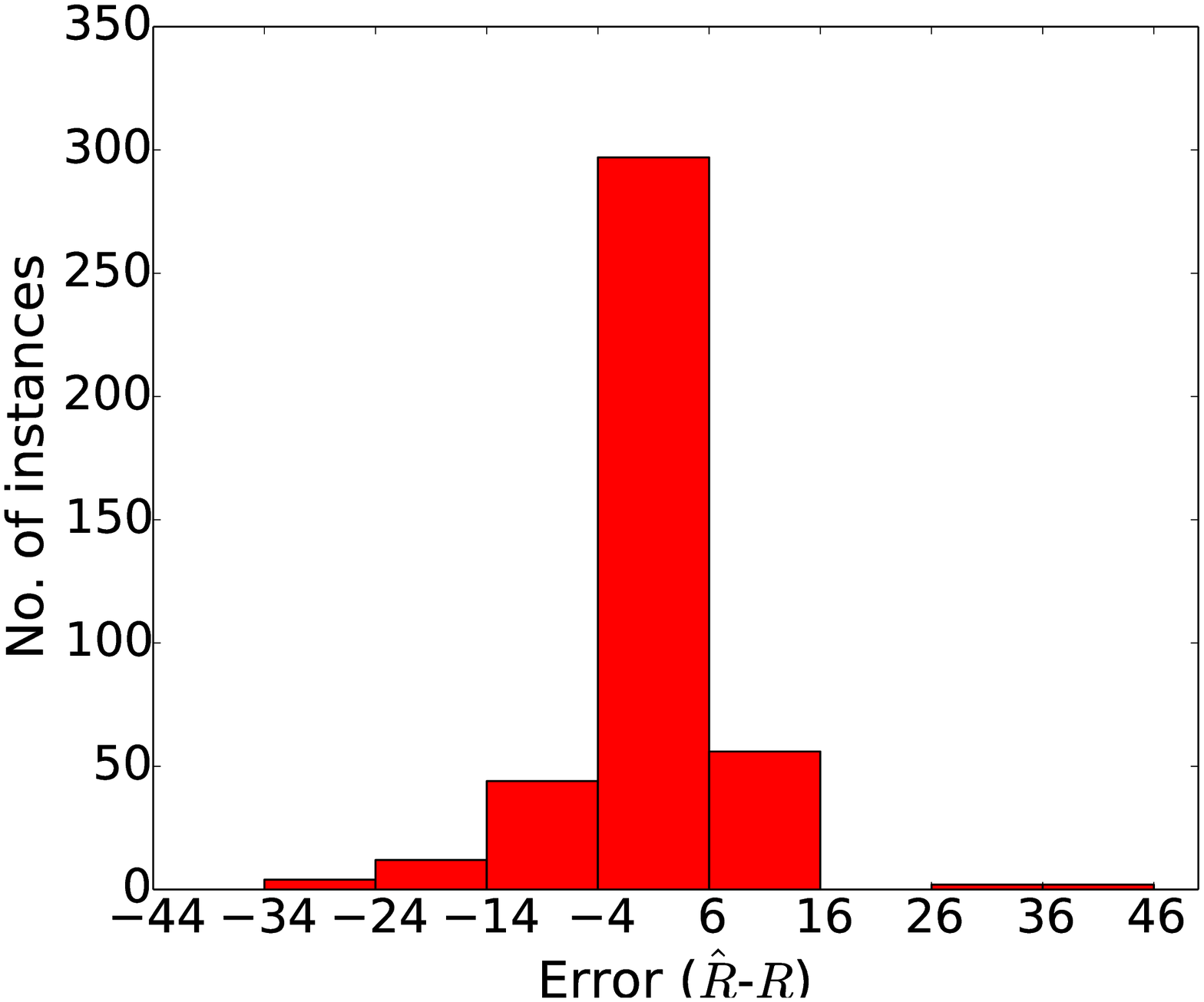}}
 \subfigure[Recon. Error Material-2]{\includegraphics[width=0.48\columnwidth]{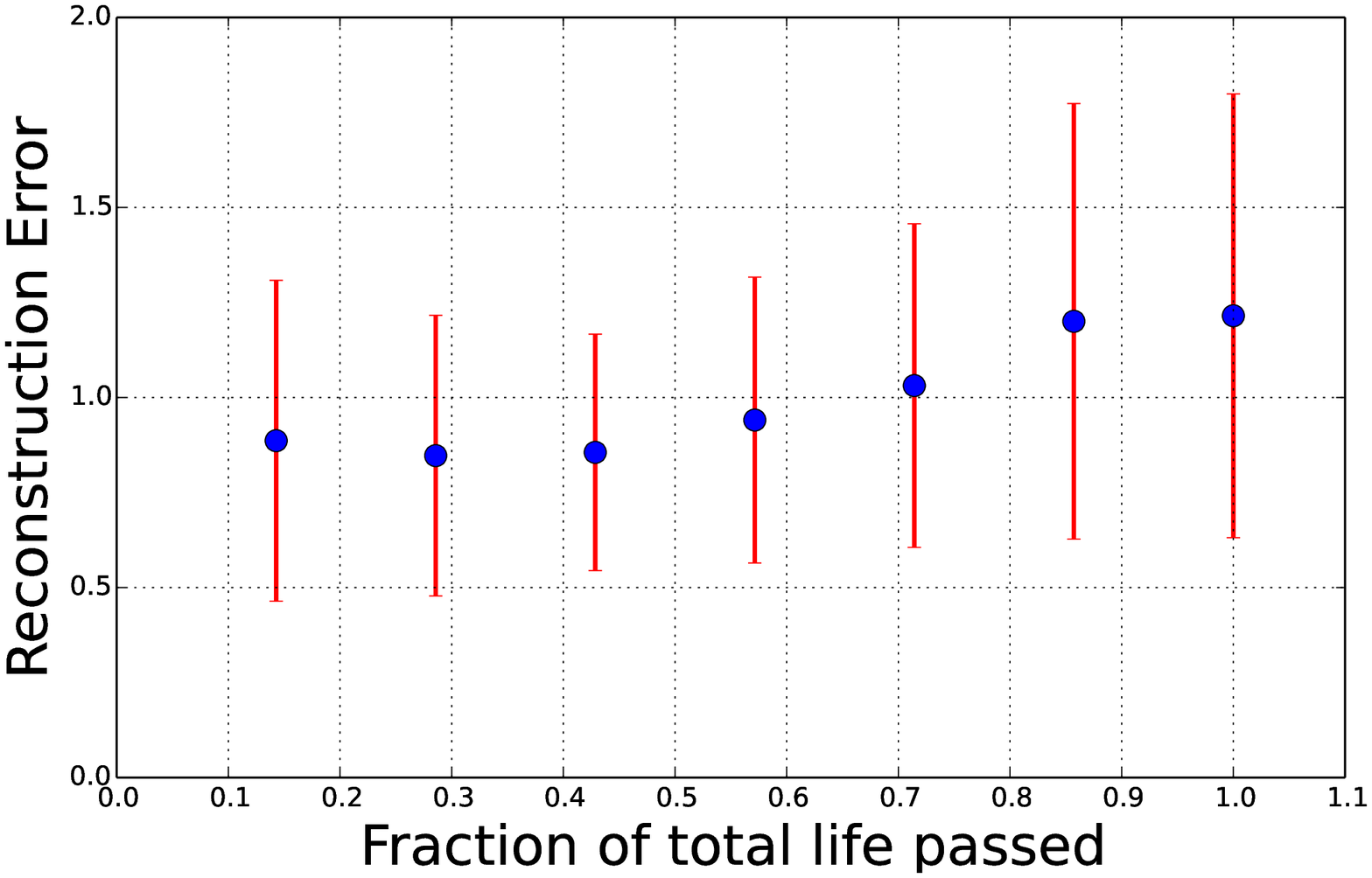}}
 \subfigure[PCA$_1$ Material-2]{\includegraphics[width=0.24\textwidth]{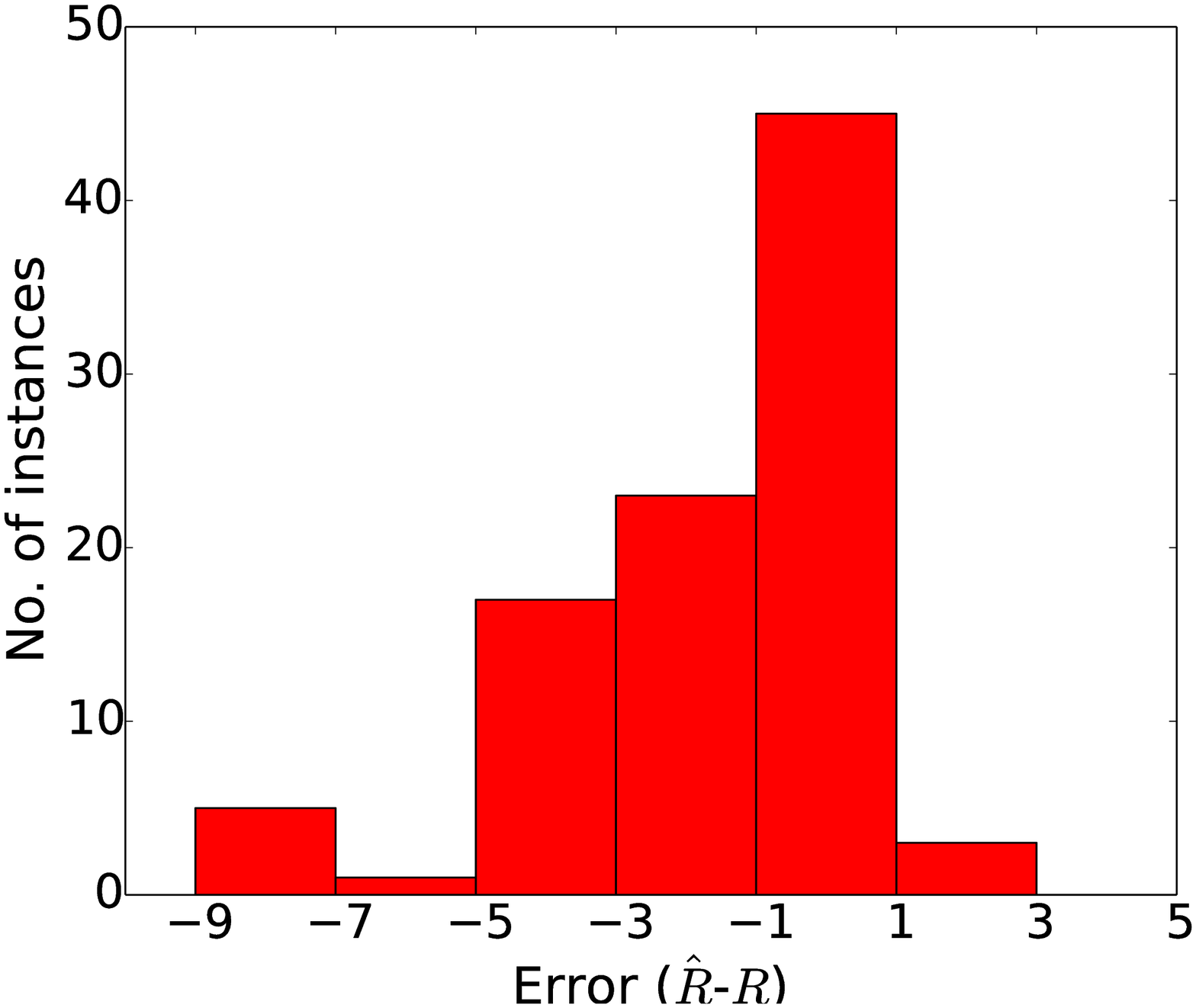}} 
 \subfigure[LR-ED$_1$ Material-2]{\includegraphics[width=0.24\textwidth]{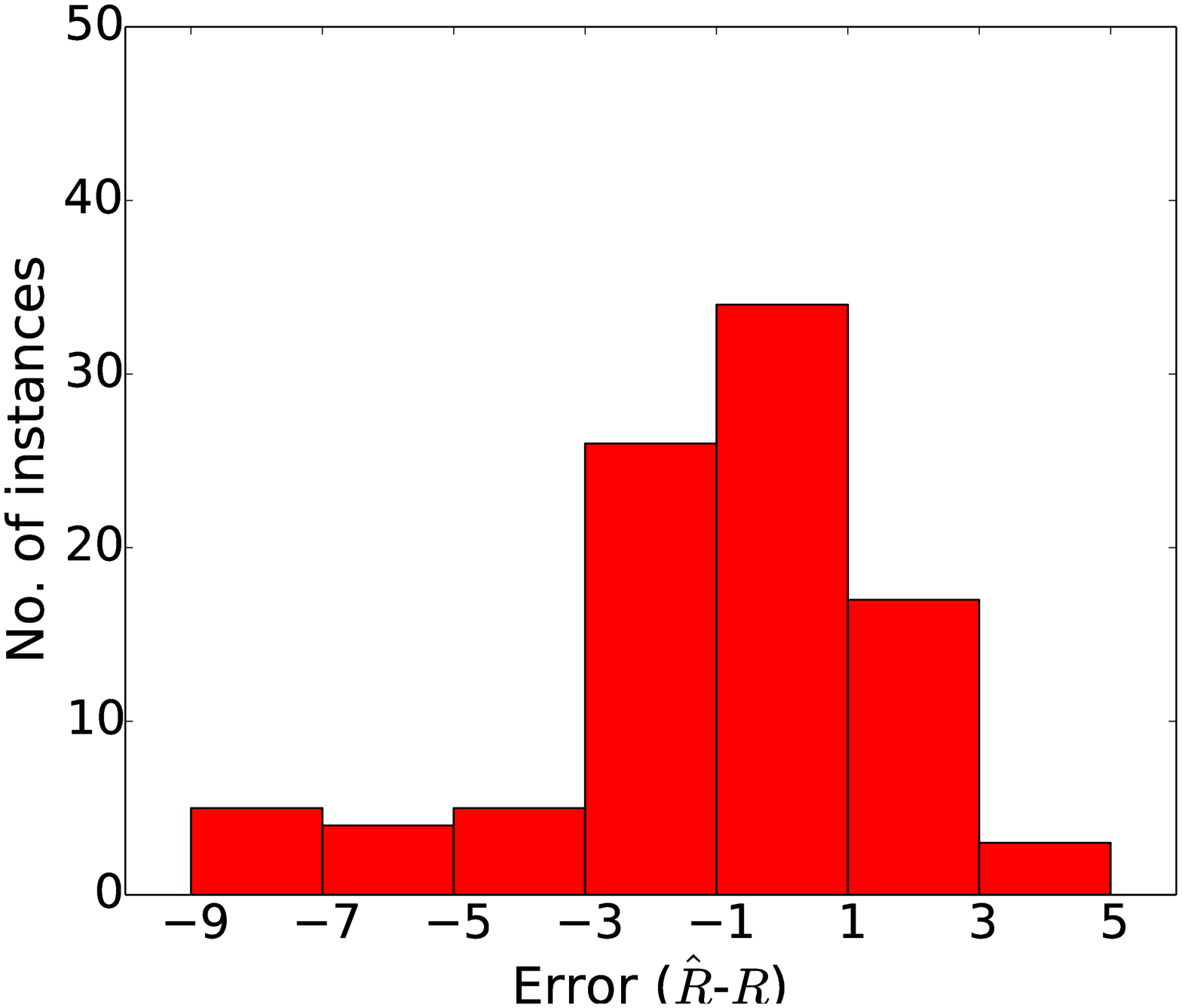}} 
 \subfigure[LR-ED$_2$ Material-2]{\includegraphics[width=0.24\textwidth]{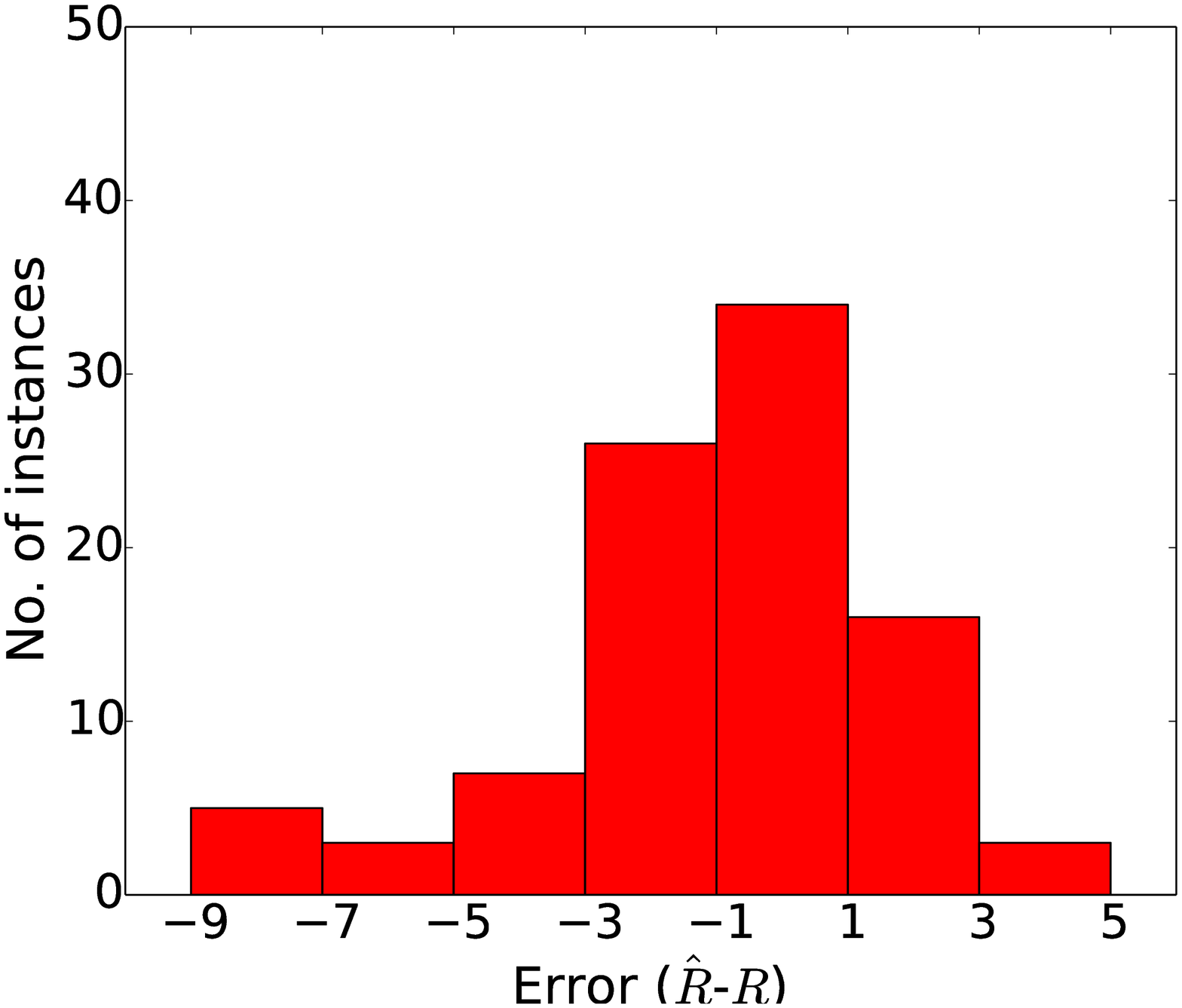}}
 \caption{\label{fig:err_hist}Milling Machine: Reconstruction errors w.r.t. cycles passed and histograms of prediction errors for milling machine dataset.}
 \vspace{-15pt}
\end{figure*}
This data set presents milling tool wear measurements from a lab experiment. Flank wear is measured for $16$ cases with each case having varying number of runs of varying durations. The wear is measured after runs but not necessarily after every run. The data contains readings for $10$ variables ($3$ operating condition variables, $6$ dependent sensors, 1 variable measuring time elapsed until completion of that run). 
\begin{figure}
 \centering
 \vspace{-5pt}
 \subfigure[Material-1]{ \vspace{-5pt}\includegraphics[scale=0.3]{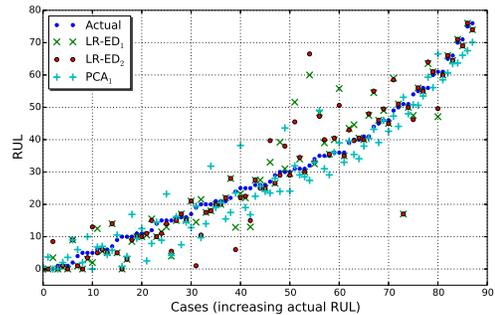}}
 \vspace{-10pt}
  \subfigure[Material-2]{ \vspace{-5pt}\includegraphics[scale=0.3]{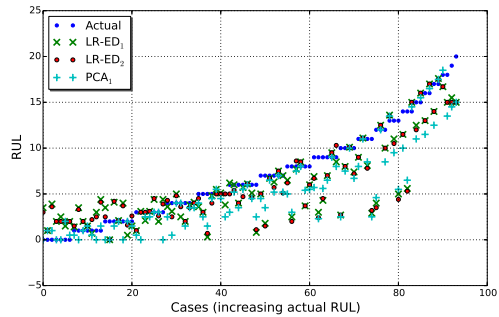}}
 \caption{Milling Machine: RUL predictions at each cycle after interpolation. Note: For material-1, estimates for every 5th cycle of all cases are shown for clarity.\label{fig:predictions-milling}}
 \vspace{-10pt}
\end{figure}
A snapshot sequence of 9000 points during a run for the $6$ dependent sensors is provided. We assume each run to represent one cycle in the life of the tool. We consider two operating regimes corresponding to the two types of material being milled, and learn a different model for each material type. There are a total of 167 runs across cases with 109 runs and 58 runs for material types 1 and 2, respectively. Case number 6 of material 2 has only one run, and hence not considered for experiments.

\subsubsection{Model learning and parameter selection}
Since number of cases is small, we use leave one out method for model learning and parameters selection. For training the LSTM-ED model, the first run of each case is considered as normal with sequence length of 9000.
An average of the reconstruction error for a run is used to the get the target HI for that run/cycle.
We consider mean and standard deviation of each run (9000 values) for the 6 sensors to obtain 2 derived sensors per sensor (similar to \cite{thesis:2datasets}). We reduce the gap between two consecutive runs, via linear interpolation, to 1 second (if it is more); as a result HI curves for each case will have a cycle of one second. The tool wear is also interpolated in the same manner and the data for each case is truncated until the point when the tool wear crosses a value of $0.45$ for the first time. The target HI from LSTM-ED for the LR model is also interpolated appropriately for learning the LR model.

The parameters obtained for the best models (based on minimum MAPE$_1$) for material-1 are $p = 1$, $\lambda = 0.025$, $\alpha = 0.98$, $\tau= 15$ for PCA$_1$, for material-2 are $p = 2$, $\lambda = 0.005$, $\alpha = 0.87$, $\tau= 13$, and $c = 45$ for LR-ED$_1$. The best results are obtained without setting any limit $R_{max}$. For both cases, $l=90$ (after downsampling by 100) s.t. the time-series for first run is used for learning the LSTM-ED model.

\subsubsection{Results and observations}
Figs. 7(a) and 7(e) show the variation of average reconstruction error from LSTM-ED w.r.t. the fraction of life passed for both materials. As shown, this error increases with amount of life passed, and hence is an appropriate indicator of health.

Figs. $7$ and $8$ show results based on every cycle of the data after interpolation, except when explicitly mentioned in the figure. The performance metrics on the original data points in the data set are summarized in Table \ref{tab:millPerf1}. We observe that the first PCA component (PCA$_1$, $p=1$) gives better results than LR-Lin and LR-Exp models with $p\geq 2$, and hence we present results for PCA$_1$ in Table \ref{tab:millPerf1}. It is to be noted that for $p=1$, all the four models LR-Lin, LR-Exp, LR-ED$_1$, and LR-ED$_2$ will give same results since all models will predict a different linearly scaled value of the first PCA component. PCA$_1$ and LR-ED$_1$ are the best models for material-1 and material-2, respectively. 
We observe that our best models perform well as depicted in histograms in Fig. 7. For the last few cycles, when actual RUL is low, an error of even 1 in RUL estimation leads to MAPE$_1$ of $100\%$. Figs. 7(b-d), 7(f-h) show the error distributions for different models for the two materials. As can be noted, most of the RUL prediction errors (around 70\%) lie in the ranges [-4, 6] and [-3, 1] for material types 1 and 2, respectively. Also, Figs. 8(a) and 8(b) show predicted and actual RULs for different models for the two materials.

\begin{table}
\centering
\resizebox{\columnwidth}{!}{
\begin{tabular}{|c|>{\centering}*{3}{p{1.2cm}|}>{\centering}*{3}{p{1.2cm}|}}\hline
& \multicolumn{3}{c|}{\textbf{Material-1}} & \multicolumn{3}{c|}{\textbf{Material-2}}\\
\hline
\textbf{Metric} & \textbf{PCA$_1$}& \textbf{LR-ED$_1$}& \textbf{LR-ED$_2$}& \textbf{PCA$_1$}& \textbf{LR-ED$_1$}&\textbf{LR-ED$_2$}\\ \hline
\textbf{MAE} & \textbf{4.2} & $\quad$4.9&$\quad$4.8& 2.1 &$\quad$\textbf{1.7}&$\quad$1.8 \\ \hline
\textbf{MSE} & \textbf{29.9} & $\quad$71.4 & $\quad$65.7 & 8.2&$\quad$\textbf{7.1}&$\quad$7.8  \\ \hline
\textbf{MAPE$_1$(\%)} &\textbf{25.4}&$\quad$26.0&$\quad$28.4& 35.7& $\quad$\textbf{31.7}&$\quad$35.5\\ \hline
\textbf{MAPE$_2$(\%)} &\textbf{9.2}&$\quad$10.6 &$\quad$10.8& 14.7& $\quad$\textbf{11.6}&$\quad$12.6\\ \hline
\end{tabular}
}
\caption{\label{tab:millPerf1}Milling Machine: Performance Comparison}
\vspace{-5pt}
\end{table}

\subsection{Pulverizer Mill Dataset}

This dataset consists of readings for $6$ sensors (such as bearing vibration, feeder speed, etc.) for over three years of operation of a pulverizer mill. The data corresponds to sensor readings taken every half hour between four consecutive scheduled maintenances $M_0$, $M_1$, $M_2$, and $M_3$, s.t. the operational period between any two maintenances is roughly one year. Each day's multivariate time-series data with length $l=48$ is considered to be one subsequence. Apart from these scheduled maintenances, maintenances are done in between whenever the mill develops any unexpected fault affecting its normal operation. The costs incurred for any of the scheduled maintenances and unexpected maintenances are available. We consider the (z-normalized) original sensor readings directly for analysis rather than computing the PCA based derived sensors. 
\begin{figure}
 \centering
\includegraphics[width=\columnwidth]{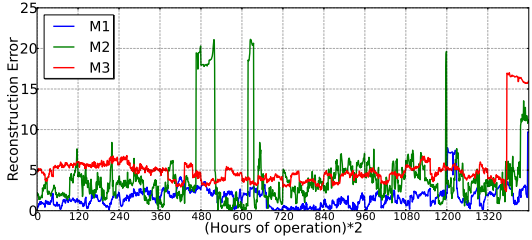}
\vspace{-14pt}
 \caption{Pulverizer Mill: Pointwise reconstruction errors for last $30$ days before maintenance.}
 \vspace{-14pt}
\end{figure}

We assume the mill to be healthy for the first $10\%$ of the days of a year between any two consecutive time-based maintenances $M_i$ and $M_{i+1}$, and use the corresponding subsequences for learning LSTM-ED models. This data is divided into training and validation sets. A different LSTM-ED model is learnt after each maintenance. The architecture with minimum average reconstruction error over a validation set is chosen as the best model. The best models learnt using data after $M_0$, $M_1$ and $M_2$ are obtained for $c=40,\:20$, and $100$, respectively. The LSTM-ED based reconstruction error for each day is z-normalized using the mean and standard deviation of the reconstruction errors over the sequences in validation set. 

From the results in Table \ref{tab:pulverizer_mill}, we observe that average reconstruction error $E$ on the last day before $M_1$ is the least, and so is the cost $C(M_1)$ incurred during $M_1$. For $M_2$ and $M_3$, $E$ as well as corresponding $C(M_i)$ are higher compared to those of $M_1$. Further, we observe that for the days when average value of reconstruction error $E>t_E$, a large fraction (>0.61) of them have a high ongoing maintenance cost $C>t_C$. The significant correlation between reconstruction error and cost incurred suggests that the LSTM-ED based reconstruction error is able to capture the health of the mill.

\begin{table}
\centering
\resizebox{\columnwidth}{!}{
\begin{tabular}{|c|c|c|c|c|c|c|} \hline
\textbf{Maint. ID}&\textbf{t$_E$}&\textbf{P(E > t$_E$)}&\textbf{t$_C$}&\textbf{P(C>t$_C$ | E>t$_E$)}&\textbf{E (last day)}&\textbf{C(M$_i$)}\\ \hline
\textbf{$M_1$} &1.50&0.25&7&0.61&2.4&92\\ \hline
\textbf{$M_2$} &1.50&0.57&7&0.84&8.0&279 \\ \hline
\textbf{$M_3$} &1.50&0.43&7&0.75&16.2&209\\ \hline
\end{tabular}
}
\caption{\label{tab:pulverizer_mill}Pulverizer Mill: Correlation between reconstruction error $E$ and maintenance cost $C(M_i)$.}
\vspace{-10pt}
\end{table}

\section{Related Work}\label{sec:Related_Work}
Similar to our idea of using reconstruction error for modeling normal behavior, variants of Bayesian Networks have been used to model the joint probability distribution over the sensor variables, and then using the evidence probability based on sensor readings as an indicator of normal behaviour (e.g. \cite{dubrawski2007applying,tobon2012cnc}). \cite{mosallam2015component} presents an unsupervised approach which does not assume a form for the HI curve and uses discrete Bayesian Filter to recursively estimate the health index values, and then use the k-NN classifier to find the most similar offline models. Gaussian Process Regression (GPR) is used to extrapolate the HI curve if the best class-probability is lesser than a threshold.

Recent review of statistical methods is available in \cite{ramasso2014performance,tsui2015prognostics,jouin2015particle}. 
We use HI Trajectory Similarity Based Prediction (TSBP) for RUL estimation similar to \cite{p:similarity_wang2008} with the key difference being that the model proposed in \cite{p:similarity_wang2008} relies on exponential assumption to learn a regression model for HI estimation, whereas our approach uses LSTM-ED based HI to learn the regression model. Similarly, \cite{p:rulclipper} proposes RULCLIPPER (RC) which to the best of our knowledge has shown the best performance on C-MAPSS when compared to various approaches evaluated using the dataset \cite{ramasso2014performance}. RC tries various combinations of features to select the best set of features and learns linear regression (LR) model over these features to predict health index which is mapped to a HI polygon, and similarity of HI polygons is used instead of univariate curve matching. The parameters of LR model are estimated by assuming exponential target HI (refer Eq. \ref{eq:expAssumption}). Another variant of TSBP \cite{thesis:tsbp} directly uses multiple PCA sensors for multi-dimensional curve matching which is used for RUL estimation, whereas we obtain a univariate HI by taking a weighted combination of the PCA sensors learnt using the unsupervised HI obtained from LSTM-ED as the target HI. 
Similarly, \cite{p:9pcMAPE,liuoptimize} learn a composite health index based on the exponential assumption.

Many variants of neural networks have been proposed for prognostics (e.g. \cite{p:deepCNNPrognostics,peng2012modified,p:rnnRUL1,p:rnnRUL2}): very recently, deep Convolutional Neural Networks have been proposed in \cite{p:deepCNNPrognostics} and shown to outperform regression methods based on Multi-Layer Perceptrons, Support Vector Regression and Relevance Vector Regression to directly estimate the RUL (see Table \ref{tab:aircraftPerf1} for performance comparison with our approach on the Turbofan Engine dataset). The approach uses deep CNNs to learn higher level abstract features from raw sensor values which are then used to directly estimate the RUL, where RUL is assumed to be constant for a fixed number of starting cycles and then assumed to decrease linearly. Similarly, Recurrent Neural Network has been used to directly estimate the RUL \cite{p:rnnRUL1} from the time-series of sensor readings. Whereas these models directly estimate the RUL, our approach first estimates health index and then uses curve matching to estimate RUL. To the best of our knowledge, our LSTM Encoder-Decoder based approach performs significantly better than Echo State Network \cite{peng2012modified} and deep CNN based approaches \cite{p:deepCNNPrognostics} (refer Table \ref{tab:aircraftPerf1}).

Reconstruction models based on denoising autoencoders \cite{marchi2015novel} have been proposed for novelty/anomaly detection but have not been evaluated for RUL estimation task.
LSTM networks have been used for anomaly/fault detection \cite{p:lstm-ad,p:lstm-ad-ecg,p:lstm-ad-ode}, where deep LSTM networks are used to learn prediction models for the normal time-series. The likelihood of the prediction error is then used as a measure of anomaly. These models predict time-series in the future and then use prediction errors to estimate health or novelty of a point. Such models rely on the the assumption that the normal time-series should be predictable, whereas LSTM-ED learns a representation from the entire sequence which is then used to reconstruct the sequence, and therefore does not depend on the predictability assumption for normal time-series.


\section{Discussion}\label{sec:Discussion}
We have proposed an unsupervised approach to estimate health index (HI) of a system from multi-sensor time-series data. The approach uses time-series instances corresponding to healthy behavior of the system to learn a reconstruction model based on LSTM Encoder-Decoder (LSTM-ED). We then show how the unsupervised HI can be used for estimating remaining useful life instead of relying on domain knowledge based degradation models. The proposed approach shows promising results overall, and in some cases performs better than models which rely on assumptions about health degradation for the Turbofan Engine and Milling Machine datasets. Whereas models relying on assumptions such as exponential health degradation cannot attune themselves to instance specific behavior, the LSTM-ED based HI uses the temporal history of sensor readings to predict the HI at a point. A case study on real-world industry dataset of a pulverizer mill shows signs of correlation between LSTM-ED based reconstruction error and the cost incurred for maintaining the mill, suggesting that LSTM-ED is able to capture the severity of fault.

\bibliographystyle{abbrv}
\bibliography{bibTeX/phm-kdd,bibTeX/icml2016,bibTeX/nips2015,bibTeX/esann}
\appendix

\section{Benchmarks on Turbofan Engine Dataset}
\begin{table}[h]
\centering
\resizebox{\columnwidth}{!}{
\begin{tabular}{|c|c|c|c|c|c|c|c|c|c|} \hline
\textbf{Approach}&\textbf{S}&\textbf{A}&\textbf{MAE}&\textbf{MSE}&\textbf{MAPE$_1$}&\textbf{MAPE$_2$}&\textbf{FPR}&\textbf{FNR}\\ \hline
\textbf{LR-ED$_2$ (proposed)}&256&\textbf{67}&\textbf{9.9}&\textbf{164}&18&\textbf{5.0}&\textbf{13}&20\\ \hline
\textbf{RULCLIPPER \cite{p:rulclipper}}&\textbf{216}&\textbf{67}&\textbf{10.0}&176&20&NR&56&44\\ \hline
\textbf{Bayesian-1 \cite{mosallam2014data}}&NR&NR&NR&NR&12&NR&NR&NR\\ \hline
\textbf{Bayesian-2 \cite{mosallam2015component}}&NR&NR&NR&NR&\textbf{11}&NR&NR&NR\\ \hline
\textbf{HI-SNR \cite{liuoptimize}}&NR&NR&NR&NR&NR&8&NR&NR\\ \hline
\textbf{ESN-KF \cite{peng2012modified}}&NR&NR&NR&4026&NR&NR&NR&NR\\ \hline
\textbf{EV-KNN \cite{ramasso2013joint}}&NR&53&NR&NR&NR&NR&36&\textbf{11}\\ \hline
\textbf{IBL \cite{khelif2014rul}}&NR&54&NR&NR&NR&NR&18&28\\ \hline
\textbf{Shapelet \cite{khelif2014rul}}&652&NR&NR&NR&NR&NR&NR&NR\\ \hline
\textbf{DeepCNN \cite{p:deepCNNPrognostics}}&1287&NR&NR&340&NR&NR&NR&NR\\ \hline

\end{tabular}
}
\caption{\label{tab:aircraftPerf1}Performance of various approaches on Turbofan Engine Data. NR: Not Reported.}
\end{table}

\end{document}